\documentclass[10 pt,journal]{IEEEtran}
\usepackage[dvips]{graphicx} % include the graphics package
\usepackage{epstopdf}
\usepackage{subfig}
\usepackage{url}
\usepackage{cite}
\usepackage{amsmath,amssymb,amsfonts}
\usepackage{algorithmic}
\usepackage{graphicx}
\usepackage{textcomp}
\usepackage{enumitem}
\usepackage{xcolor}
\usepackage{csquotes}

\def\BibTeX{{\rm B\kern-.05em{\sc i\kern-.025em b}\kern-.08em
    T\kern-.1667em\lower.7ex\hbox{E}\kern-.125emX}}
\newcommand{\x}{\mbox{$\mathbf{x}$}}

\newcommand{\cvec}{\mbox{$\mathbf{c}$}}
\newcommand{\chat}{\mbox{$\hat{\mathbf{c}}$}}
\newcommand{\y}{\mbox{$\mathbf{y}$}}
\newcommand{\z}{\mbox{$\mathbf{z}$}}
\newcommand{\zhat}{\mbox{$\hat{\mathbf{z}}$}}
\newcommand{\bb}{\mbox{$\mathbf{b}$}}
\newcommand{\bhat}{\mbox{$\hat{\mathbf{b}}$}}

\newcommand{\D}{\mbox{$\mathbf{D}$}}
\newcommand{\U}{\mbox{$\mathbf{U}$}}

\newcommand{\I}{\mbox{$\mathbf{I}$}}

\newcommand{\Mu}{\mbox{$\mathbf{m}$}}
\newcommand{\Muhat}{\mbox{$\hat{\mathbf{m}}$}}
\newcommand{\Sig}{\mbox{$\mathbf{\Sigma}$}}
\newcommand{\Sighat}{\mbox{$\hat{\mathbf{\Sigma}}$}}

\begin{document}
\title{A Doubly Regularized Linear Discriminant Analysis Classifier with Automatic Parameter Selection}
\author
{
Alam~Zaib,~Tarig~Ballal,~\IEEEmembership{Member,~IEEE},~Shahid~Khattak ~and ~Tareq~Y.~Al-Naffouri, ~\IEEEmembership{Senior Member,~IEEE}

\thanks{A. Zaib and S. Khattak are with the Department of Electrical \& Computer Engineering, COMSATS University Islamabad, Abbottabad Campus, Pakistan, \mbox{E-mails: \{alamzaib, skhattak\}@cuiatd.edu.pk}. T. Ballal and T. Y. Al-Naffouri are with the Department of Electrical Engineering, King Abdullah University of Science and Technology (KAUST), Thuwal 23955-6900, Saudi Arabia, \mbox{E-mails: \{tarig.ahmed, tareq.alnaffouri\}@kaust.edu.sa}.}
}

\maketitle

\begin{abstract}
Linear discriminant analysis (LDA) based classifiers tend to falter in many practical settings where the training data size is smaller than, or comparable to, the number of features. As a remedy, different regularized LDA (RLDA) methods have been proposed. These methods may still perform poorly depending on the size and quality of the available training data. \textcolor{black}{In particular, the test data deviation from the training data model, for example, due to noise contamination, can cause severe performance degradation. Moreover, these methods commit further to the Gaussian assumption (upon which LDA is established) to tune their regularization parameters, which may compromise accuracy when dealing with real data.} To address these issues, we propose a doubly regularized LDA classifier that we denote as R2LDA. In the proposed R2LDA approach, the RLDA score function is converted into an inner product of two vectors. By substituting the expressions of the regularized estimators of these vectors, we obtain the R2LDA score function that involves two regularization parameters. To set the values of these parameters, we adopt three existing regularization techniques; the constrained perturbation regularization approach (COPRA), the bounded perturbation regularization (BPR) algorithm, and the generalized cross-validation (GCV) method. These methods are used to tune the regularization parameters based on linear estimation models, with the sample covariance matrix's square root being the linear operator. Results obtained from both synthetic and real data demonstrate the consistency and effectiveness of the proposed R2LDA approach, especially in scenarios involving test data contaminated with noise that is not observed during the training phase.
\end{abstract}


\section{Introduction}
\label{sec:ntro}
The idea of linear discriminant analysis (LDA) was originally conceived by R. A. Fisher \cite{fisher_lda} and is based on the assumption that the data follows a Gaussian distribution with a common class covariance matrix. Owing to its simplicity, LDA has been successfully applied to various classification and recognition tasks such as detection \cite{detect_lda}, speech recognition \cite{speech_lda}, cancer genomics \cite{cancer_lda, cancer_lda2} and face recognition \cite{image_lda} to mention a few. In addition, LDA is a classical tool for feature extraction \cite{Liu2020}.

The performance of LDA-based classifiers depends heavily on accurate estimation of the class statistics, namely, the sample covariance matrix and class mean vectors. These statistics can be estimated with fairly high accuracy when the number of available samples is large compared to the data dimensionality. In practical high-dimensional data settings, the challenge is to cope with a limited number of available samples. In this case, the sample covariance estimates become highly perturbed and ill-conditioned resulting in severe performance degradation. To alleviate this problem, the sample covariance matrix is replaced with a regularized or ridge covariance matrix \cite{rlda1}, giving the name regularized LDA (RLDA). The values of the regularization parameters ultimately dictate the performance of RLDA classifiers. Hence, it is essential to judiciously tune the regularization parameters' values to reap the full benefit of the regularization process. Towards this end, various regularization techniques have been proposed. For example, cross-validation \cite{cross_valid} has been one of the classical techniques for estimating the ridge parameter as evidenced in \cite{cv1, cv2, cv3, cancer_lda2, cv4}.

An optimal regularization method that minimizes the asymptotic classification error is derived in \cite{zollanvari1, zollanvari2}. The method is based on recent results from random matrix theory. In \cite{khalil, Elkhalil2020}, the method of \cite{zollanvari1, zollanvari2} is extended to a more general class of discriminant analysis based classifiers, with LDA obtained as a special case. In \cite{impLDA_spawc}, \cite{Sifaou2020}, improved RLDA classifiers are proposed, with the required parameters given in closed forms. These classifiers are designed for spiked-model covariance structures. Nevertheless, the authors demonstrate their usefulness when the data is generated from other (non-spiked) models.

In all the above-mentioned RLDA approaches, a regularization parameter is tuned based only on the data available in the training phase. Such a regularization parameter may produce satisfactory results when the test data follows the exact model of the training data. In some practical situations, it occurs that the test data deviates from the training data model. For example, the training data and the test data might represent measurements obtained from non-identical devices. In such a case, the value of the regularization parameter computed during the training phase may no longer be adequate, let alone be optimal. Consequently, the above-mentioned approaches' performance might deteriorate significantly. \textcolor{black}{Moreover, these methods use the Gaussian assumption of the underlying data distribution for finding the value of the regularization parameter. This assumption may not hold in practical settings, e.g., with real data. Even though the Gaussian assumption is essential in deriving the basic LDA, excessive reliance on the assumption may eventually compromise the RLDA classifier's performance.} To tackle these issues, we propose a new approach to regularized LDA classification. Focusing on binary classification, this paper develops a doubly regularized LDA (R2LDA) classifier by expressing the LDA score function as an inner product of two vectors that are linearly related to the mean vectors and the data covariance matrix. Regularized estimators are used to obtain the values of the two vectors and the value of the score function. The regularization parameter used in the estimation of one of the two vectors is tuned based on the current sample of the test data, hence providing robustness against any irregularities in the test data.

We summarize our main innovations and the most prominent features of the proposed R2LDA approach as follows:
 \begin{enumerate} [label=(\alph*)]
   \item We deviate from the classical covariance matrix estimation approach to RLDA, where the focus is to obtain a regularized linear estimator of the data covariance matrix. Instead, we reformulate the problem as a vector estimation problem. We apply regularization to estimate two vector quantities. This implicitly results in a regularized \emph{nonlinear} estimator of the data covariance matrix.
   \item R2LDA is designed not only to cope with the insufficiency of the training data but also with perturbations in the test data that are not observed during training. This is achieved by adjusting two regularization parameters independently; one is computed based only on the training data, and another is dynamically tuned to the test data sample. This is to be contrasted with existing approaches that compute their regularization parameters based solely on the training data.
   \item We automate the regularization parameter selection process based on existing methods that are well suited to the task. We theoretically motivate the main approaches adopted to tune the regularization parameters.
   \item The regularization parameter selection approach is agnostic to the underlying distribution of the data contrary to \cite{zollanvari2, khalil, impLDA_spawc}, which rely on the Gaussian assumption. Even though the Gaussian assumption is embedded in LDA, further commitment to Gaussianity in the regularization parameter tuning process might impede classification performance, especially with real data.
 \end{enumerate}

\subsection{Notations}
\textcolor{black}{Throughout this paper, we use non-bold letters to denote scalars (e.g., $W$), boldface lowercase letters to denote column vectors (e.g., $\x$), and boldface uppercase letters to denote matrices (e.g., $\mathbf{H}$). The notation $\mathbf{I}_p$ denotes an identity matrix of dimension $p$, and $\mathbf{0}_{p_1\times p_2}$ represents a $p_1\times p_2$ matrix with all zero elements. We use $\text{tr}(.)$ and $(.)^{\rm T}$ to denote the matrix trace and matrix/vector transpose operations, respectively. The notation $\hat{x}$ indicates an estimate of the variable $x$. The set of real numbers is denoted by $\mathbb{R}$ and the $l_2$ norm of a vector is denoted by $\|.\|_2$. The probability density function and the statistical expectation of a random variable $x$ are denoted by ${\rm P}(x)$ and $\mathbb{E}(x)$, respectively. The symbol $\approx$ stands for \enquote{approximately equivalent to,} while $:=$ means \enquote{defined to be equal to}. Finally, \enquote{s.t.} is an abbreviation for \enquote{subject to.}}

The remainder of this paper is organized as follows. In Section~\ref{sec:lda}, we present a concise overview of regularized LDA classification. In Section~\ref{sec:rrlda}, we present our proposed R2LDA approach, along with three regularization parameter selection methods. Performance evaluation of the proposed approach and comparisons with existing techniques are presented in Section~\ref{sec:sim}. We close this paper by making a concluding remark in Section~\ref{sec:sim}.

\section{RLDA Classification}
\label{sec:lda}
We consider the binary classification problem of assigning a multivariate observation vector $\x\in\mathbb{R}^{p\times 1}$ to one of two classes $\mathcal{C}_i,i\! =\! 0, 1$. Let $\pi_i$ be the prior probability that $\x$ belongs to the class $\mathcal{C}_i$, and assume that the class conditional densities ${\rm P}\left(\x | \x\in \mathcal{C}_i \right), i\! =\! 0, 1$, are Gaussian with mean vectors $\Mu_i\in \mathbb{R}^{p\times 1}$ and positive semidefinite covariance matrices $\Sig_i\in \mathbb{R}^{p\times p}$.

LDA employs the Bayesian discriminant rule, which assigns $\x$ to the class with the maximum posterior probability. Let $\mathcal{S}_0\! =\! \{\x_l\}_{l=0}^{n_0}$ and $\mathcal{S}_1 \!=\! \{\x_l\}_{l=n_0+1}^{n_0+n_1}$ represent the available training samples pertaining to the two classes, where $n_i$ is the number of training samples for class $\mathcal{C}_i$ and $n \!=\! n_0 \!+\! n_1$ is the total number of training samples. The LDA score function reads \cite{Anderson1951}
\begin{equation}\label{eq:LDAhat}
  W^{\rm {LDA}}(\x) = \left(\x-\frac{\Muhat_0+\Muhat_1}{2} \right)^\mathrm{T}\Sighat^{-1}\left(\Muhat_0-\Muhat_1 \right).
\end{equation}
The unbiased mean vector estimates $\Muhat_i$, and the pooled sample covariance matrix $\Sighat$, are computed according to
\begin{equation}\label{eq:MuSig_est}
 \Muhat_i = \frac{1}{n_i}\sum_{l\in \mathcal{S}_i}\x_l, \:\:\:
 \Sighat = \frac{(n_0-1)\Sighat_0+(n_1-1)\Sighat_1}{n_0+n_1+1},
\end{equation}
where the sample covariance matrices $\Sighat_i$ are computed using
\begin{equation}\label{eq:Sighat}
 \Sighat_i = \frac{1}{n_i-1}\sum_{l\in \mathcal{S}_i}(\x_l-\Muhat_i)(\x_l-\Muhat_i)^{\rm T}.
\end{equation}
The class assignment rule for $\x$ is as follows:
\begin{equation}\label{eq:LDArule}
 \x \in \left\{ \begin{array}{lr}
        \mathcal{C}_0,\:\:  \text{if}\: W(\mathbf{x})> \log(\pi_1/\pi_0); \\
        \mathcal{C}_1, \:\: \text{otherwise}.
        \end{array} \right.
\end{equation}

A major source of error in the above formulation is the inversion of the sample covariance matrix $\Sighat$. In many practical setups where $n$ is comparable to $p$, $\Sighat$ becomes ill-conditioned, or even singular. To circumvent this issue, $\Sighat^{-1}$ in (\ref{eq:LDAhat}) is replaced with a regularized estimator. Typically, $\mathbf{H} \!=\! (\I_p \!+\! \gamma\Sighat)^{-1}$ is used, where $\gamma \in \mathbb{R}^+$ is a regularization parameter and $\I_p$ is the identity matrix of dimension $p$. This replacement results in the RLDA score function \cite{zollanvari2, zollanvari1}
\begin{equation}\label{eq:RLDA}
  W^{\rm {RLDA}}(\x) = \left(\x-\frac{\Muhat_0+\Muhat_1}{2} \right)^\mathrm{T}\mathbf{H}\left(\Muhat_0-\Muhat_1 \right).
\end{equation}

In this work, we apply a different regularization form to (\ref{eq:LDAhat}). In the proposed regularized LDA classifier, we employ two separate regularization operations to account for the deficiency in the training data. The proposed approach also improves the classifier's robustness to error contributions that are present only in the test data.

\section{The proposed R2LDA classification Approach}
\label{sec:rrlda}
Many existing RLDA techniques are based on (\ref{eq:RLDA}), with $\mathbf{H}$ estimated by selecting the regularization parameter $\gamma$ using only the training data. This makes these techniques vulnerable to errors in the test data. To address this issue, we express the LDA score function (\ref{eq:LDAhat}) as
\begin{equation}\label{eq:RRLDAhat2}
  W^{\rm {LDA}}(\x) = (\x')^{\rm T}\Sighat^{-\frac{1}{2}}\Sighat^{-\frac{1}{2}}\Muhat^-
  = \z^{\rm T}\bb ,
\end{equation}
where $\x' \!:=\! \x \!-\! \frac{1}{2}\Muhat^+$, $\Muhat^+ \!:=\! \Muhat_0\!+\!\Muhat_1$, $\Muhat^- \!:=\! \Muhat_0 \!-\! \Muhat_1$,
$\z:=  \Sighat^{-\frac{1}{2}}\x'$, and $\bb:= \Sighat^{-\frac{1}{2}} \Muhat^-$. Based on the last two definitions, our proposed R2LDA method aims to obtain regularized estimates of $\z$ and $\bb$ to improve the computation of the score function (\ref{eq:RRLDAhat2}). To this end, we utilize the linear models
%\noindent\begin{tabularx}{\linewidth}{@{}XX@{}}
\begin{equation}\label{eq:xprime}
  \x' = \Sighat^{\frac{1}{2}}\z + \mathbf{v}_x,
\end{equation}
\begin{equation}\label{eq:Muhatminus}
  \Muhat^- = \Sighat^{\frac{1}{2}}\bb + \mathbf{v}_m,
\end{equation}
where $\mathbf{v}_x$ and $\mathbf{v}_m$ are additive noise vectors. These noise vectors can be interpreted as the contribution of the errors in estimating the mean vectors. In addition, $\mathbf{v}_x$ can also be used to absorb any noise contributions that occur in the test data vector $\mathbf{x}$. Each of (\ref{eq:xprime}) and (\ref{eq:Muhatminus}) can be represented by the linear model
\begin{equation}\label{eq:ySig}
  \y=\Sighat^{\frac{1}{2}}\mathbf{c}+\mathbf{v},
\end{equation}
where (\ref{eq:xprime}) or (\ref{eq:Muhatminus}) can be obtained by setting $\{\y=\x'$, $\mathbf{c} = \mathbf{z}$, $\mathbf{v} = \mathbf{v}_x$\}, or $\{\y=\Muhat^-$, $\mathbf{c} = \mathbf{b}$,$\mathbf{v} = \mathbf{v}_m$\}, respectively.

Focusing on (\ref{eq:ySig}), regularization methods, commonly named ridge regression or Tikhonov regularization \cite{Tikhonov1,Tikhonov2, Tikhonov3}, can be applied to obtain a stabilized estimate of $\mathbf{c}$. This estimate can be expressed in a closed form as \cite{ridge3}
\begin{equation}\label{eq:chat}
  \hat{\mathbf{c}} =(\Sighat+\gamma\I_p)^{-1}\Sighat^{\frac{1}{2}}\y.
\end{equation}

Based on (\ref{eq:chat}), we can estimate $\z$ and $\bb$ and substitute the results in (\ref{eq:RRLDAhat2}) to obtain the R2LDA score function in the form
\begin{align}\label{eq:RRLDAhat3}
  W^{\rm {R2LDA}}(\x) &= \zhat^{\rm T}\,\bhat \nonumber\\
  = (\x')^{\rm T}\U\D^2&\left(\D^2+\gamma_z\I_p \right)^{-1}\left(\D^2+\gamma_b\I_p \right)^{-1}\U^{\rm T}\Muhat^- ,
\end{align}
where $\gamma_z\in\mathbb{R}^+$ and $\gamma_b\in\mathbb{R}^+$ are the regularization parameters associated with the linear models (\ref{eq:xprime}) and (\ref{eq:Muhatminus}), respectively. The second equality in (\ref{eq:RRLDAhat3}) follows directly from substituting (in (\ref{eq:chat})) the eigenvalue decomposition (EVD) $\Sighat \!=\! \U\D^2\U^{\rm T}$, where $\U$ is the matrix of eigenvectors and $\D^2$ is the diagonal matrix of eigenvalues of $\Sighat$.

Now, it only remains to set the values of the regularization parameters $\gamma_z$ and $\gamma_b$, which will be discussed in the following subsections. \\

\newtheorem{remark}{Remark}
\begin{remark}
\label{r1}
Compared to the conventional RLDA score function (\ref{eq:RLDA}), the new formulation (\ref{eq:RRLDAhat3}) involves two regularization operations. Note that the estimation of the class mean vectors $\Mu_i$ results in perturbations in both $\Muhat^{-}$ and $\x'$. Besides, $\x'$ also has errors coming from the test data. By carrying out two independent estimations to obtain regularized estimates of $\z$ and $\bb$ (see (\ref{eq:RRLDAhat2})), we can optimize the choice of two different regularization parameters to cope with the different perturbations in $\x'$ and $\Muhat^{-}$. This is a key advantage of the proposed R2LDA method over the classical RLDA based on (\ref{eq:RLDA}) that employs a single regularization operation based only on the training data.
\end{remark}

\subsection{Regularization Parameter Selection}
\label{subsec:reg}
Several methods have been proposed in the literature for selecting the regularization parameter $\gamma$ required in (\ref{eq:chat}), e.g., \cite{lcurve, book_gcv, book_quasi, quasi}, to mention a few. These methods are based on different criteria, which results in different regularization parameter values (see \cite{regComp}).

In this work, we pursue three regularization methods; the constrained perturbation regularization approach (COPRA) \cite{copra}, bounded perturbation regularization (BPR) \cite{bpr}, and the generalized cross-validation (GCV) \cite{book_gcv}. The choice of COPRA and BPR is motivated by the fact that these algorithms are designed to optimize the \emph{mean squared error} of a vector estimation. Also, these two methods are based on a very relevant model to the setup under consideration. \textcolor{black}{As will be shown subsequently, BPR is a special case of COPRA}. On the other hand, cross-validation, \textcolor{black}{a method based on a totally different concept compared to BPR and COPRA,} is a widely adopted heuristic technique that has shown immense success in machine-learning applications.

 Next, we provide details on the three selected regularization methods and how they can be combined with R2LDA.

\subsection{The Constrained Perturbation Regularization Algorithm (COPRA)}
\label{subsec:copra}
To simplify the derivations, we make the following assumptions on the model (\ref{eq:ySig}):
\begin{enumerate}
  \item The noise vector $\mathbf{v}$ has zero mean and an unknown covariance matrix $\sigma_v^2 \I_p $.
  \item The unknown random vector $\mathbf{c}$ is zero mean with an unknown positive semidefinite diagonal covariance matrix $\Sig_{\cvec\cvec}$.
  \item The vectors $\mathbf{v}$ and $\mathbf{c}$ are mutually independent.
\end{enumerate}

COPRA is based on the principle of introducing an artificial perturbation in a linear model to improve the singular-value structure of the resulting model matrix. For the linear model in (\ref{eq:ySig}), $\Sighat^{\frac{1}{2}}$ is replaced by a perturbed version to obtain the model
\begin{equation}
\label{eq:yDel}
  \y \approx \left(\Sighat^{\frac{1}{2}} + \mathbf{\Delta}\right)\cvec+\mathbf{v},
\end{equation}
where $\mathbf{\Delta} \!\in\! \mathbb{R}^{p\times p}$ is an unknown perturbation matrix which is norm bounded by a positive quantity $\lambda$, i.e., $\|\mathbf{\Delta} \|_2 \!\le\! \lambda$. The original method in \cite{copra} utilizes the perturbation $\Delta$ to stabilize the estimation of $\mathbf{c}$ based on the model (\ref{eq:ySig}). However, in this specific application, $\Delta$ can be viewed as a genuine \emph{uncertainty} in the model due to the noisy nature of $\Sighat^{\frac{1}{2}}$. In other words, \eqref{eq:yDel} is the natural model for our vector estimation problem. These two different interpretations of $\Delta$ in \eqref{eq:yDel} yield identical estimators of the vector $\mathbf{c}$ (i.e., the same value of the regularization parameter in \eqref{eq:chat}). This makes COPRA an excellent candidate for computing the regularization parameters for R2LDA.

To obtain an estimate of $\mathbf{c}$, we consider the minimization of the worst-case residual error. Namely, we pursue the following optimization:
\begin{equation}\label{eq:yError}
 \underset{\hat{\mathbf{c}}}{\min} \,\,\underset{\mathbf{\Delta}}{\max}\, \Big\|\y - \left(\Sighat^{\frac{1}{2}}+ \mathbf{\Delta} \right) \hat{\mathbf{c}} \Big\|_2, \,\,\text{s.t.} \,\, \|\mathbf{\Delta}\|_2 \leq \lambda.
\end{equation}
Interestingly, as shown in \cite{ASayed,copra,ballal_icaasp}, the min-max problem (\ref{eq:yError}) can be converted to a minimization problem whose solution is given by (\ref{eq:chat}), with the additional constraint
\begin{equation}
\label{eq:gamma-lambda}
 \gamma\|\chat \|_2 = \lambda\Big\|\y - \Sighat^{\frac{1}{2}}\chat \Big\|_2 .
\end{equation}
Based on \eqref{eq:gamma-lambda}, we observe that the solution of (\ref{eq:yError}) depends on the bound $\lambda$ (in addition to the other system parameters) and is agnostic to the structure of the perturbation matrix $\mathbf{\Delta}$.

Now, we can substitute (\ref{eq:chat}) and the EVD of $\Sighat$ in (\ref{eq:gamma-lambda}) and manipulate to obtain
\begin{equation}\label{eq:lamdamin}
\lambda^{2} = \frac{\text{tr}\left( \left(\D^{2} + \gamma \I_p \right)^{-2} \U^{\rm T}\, \y \y^{\rm T}\, \U  \right)}{\text{tr}\left( \D^{2} \left(\D^{2} + \gamma \I_p \right)^{-2}
\U^{\rm T}\, \y \y^{\rm T}\, \U  \right)}.
\end{equation}
where $\text{tr}(.)$ is the matrix trace operation. Since $\lambda$ in (\ref{eq:lamdamin}) is stochastic in nature (due to the involvement of $\y$), we consider a value of $\lambda$ that would represent the average case. To this end, we replace $\y\y^{\rm T}$ with its expected value $\mathbb{E}(\y\y^{\rm T})$, which can be written based on (\ref{eq:ySig}) in the following form:
\begin{equation}\label{eq:Eyy}
  \mathbb{E}(\y\y^{\rm T})= \U \D \U^{\rm T}\Sig_{\cvec\cvec} \U \D \U^{\rm T} + \sigma_v^{2} \I_p.
\end{equation}
Owing to the ill-conditioning of $\Sighat$, it is likely that some of its eigenvalues are very close, or even equal, to zero. Therefore, the EVD of $\Sighat$ can be written in the form
\begin{equation}
\label{eq:EVD2}
 \Sighat = \left[ \U_1\:\:\U_2 \right] \left[ \begin{array}{cc}
                                                     \D_1^2 & \mathbf{0}_{p_1\times p_2} \\
                                                     \mathbf{0}_{p_2\times p_1} & \D_2^2 \\                                     \end{array}
                                                 \right] \left[
                                                           \begin{array}{c}
                                                             \U_1^{\rm T} \\
                                                             \U_2^{\rm T} \\
                                                           \end{array}
                                                         \right] \simeq \U_1\D_1^2\U_1^{\rm T},
\end{equation}
where $\D_1$ and $\D_2$ are diagonal matrices containing the $p_1$ most significant and $p_2 = p - p_1$ least significant eigenvalues, respectively. A threshold based approach to find the point of this partitioning is recommended in \cite{copra}. However, a simple and intuitive rule is used here to determine the value of $p_1$ as the smaller value of $p$ (the number of features) and $n$ (the number of training samples), i.e., $p_1 = \min(n, p)$. The main purpose of (\ref{eq:EVD2}) is to improve numerical stability by removing extremely small eigenvalues.

Now, we substitute (\ref{eq:Eyy}) and (\ref{eq:EVD2}) in (\ref{eq:lamdamin}) and manipulate to obtain (\ref{eq:lambdamin4}) (as shown on the top of the following page).
\begin{figure*}[t]
\setcounter{equation}{17}
\begingroup
    \fontsize{9.25pt}{9.25pt}
\begin{equation}\label{eq:lambdamin4}
    \lambda^2 \left(\text{tr}\left(\left(\D_{1}^2 + \gamma \I_{p_1} \right)^{-2} \left(\D_{1}^2 + \frac{p_1 \sigma_v^{2}}{\text{tr}\left(\Sig_{\cvec\cvec}\right)} \I_{p_1} \right)  \right) + \frac{(p-p_1) p_1 \sigma_v^{2}}{\gamma^2{\text{tr}\left(\Sig_{\cvec\cvec}\right)}} \right)\simeq
     \,\, {\text{tr}\left( \D_{1}^2 \left(\D_{1}^2 + \gamma \I_{p_1} \right)^{-2} \left(\D_{1}^2 + \frac{p_1 \sigma_v^{2}}{\text{tr}\left(\Sig_{\cvec\cvec}\right)} \I_{p_1} \right)  \right)}
\end{equation}
\endgroup
\vspace{-10pt}
\end{figure*}
Next, we proceed to eliminate $\sigma_v$ and $\Sig_{\cvec\cvec}$ from (\ref{eq:lambdamin4}) by using the \emph{mean squared error} (MSE) as a performance criterion. The MSE of the RLS estimator (\ref{eq:chat}) can be written as \cite{ridge3}
\begin{align}\label{eq:MSE2}
\text{MSE} \!=\!  \text{tr}& \left( \mathbb{E}\left( (\mathbf{c} - \hat{\mathbf{c}}) (\mathbf{c} - \hat{\mathbf{c}})^{\rm T}  \right) \right)
 \!=\! \sigma_v^{2} \text{tr}\left(\D^{2} \left(\D^{2}\! +\! \gamma\I_p \right)^{-2} \right) \nonumber \\
& + \gamma^{2} \text{tr}\left( \left(\D^{2} \!+\! \gamma\I_p \right)^{-2}\U^{\rm T}\Sig_{\mathbf{c}\mathbf{c}}\U \right).
\end{align}

By differentiating (\ref{eq:MSE2}), the regularization parameter $\gamma$ that minimizes the MSE can be obtained using
\begin{equation}\label{eq:gammaoptapprox}
\frac{ \partial \left(\text{MSE}\right)}{\partial \  \gamma}=0 \implies \gamma = \frac{p\, \sigma_v^{2}}{\text{tr}\left(\Sig_{\cvec\cvec}\right)}.
\end{equation}

By substituting (\ref{eq:gammaoptapprox}) in (\ref{eq:lambdamin4}), we obtain (\ref{eq:lambdamin5}), which shows a bound $\lambda$ that does not depend on the statistics of $\cvec$ or those of the noise. Note that the derivations of (\ref{eq:Eyy}) and (\ref{eq:lambdamin4}) require Assumptions~1--3 to be satisfied--otherwise, these results will hold only in an approximation way.

Ultimately, by using (\ref{eq:lambdamin5}), we can eliminate $\lambda$ from (\ref{eq:lamdamin}) to obtain (\ref{eq:secularEq2}), where $\mathbf{d}:=\U^{\rm T}\y$. Equation~(\ref{eq:secularEq2}), which is nonlinear in $\gamma$, can be solved by using Newton's method \cite{newton-method} to obtain the optimal value of $\gamma$. The iterations should be initialized from a positive initial guess close to zero to avoid missing the positive root, as explained in \cite{copra}.

\begin{figure*}[t]
\setcounter{equation}{20}
\begingroup
    \fontsize{9.25pt}{9.25pt}
\begin{equation}\label{eq:lambdamin5}
\lambda^2 \left(\text{tr}\left( \left(\D_{1}^2 + \gamma \I_{p_1} \right)^{-2}  \left(\frac{p}{p_1}\D_{1}^2 + \gamma \I_{p_1} \right) \right) +\frac{(p-p_1)}{\gamma}\right) \simeq
\text{tr}\left( \D_{1}^2 \left(\D_{1}^2 + \gamma \I_{p_1} \right)^{-2} \left(\frac{p}{p_1}\D_{1}^2 + \gamma \I_{p_1} \right)  \right)
\end{equation}
\endgroup
\vspace{-2pt}
\begingroup
    \fontsize{9.25pt}{9.25pt}
\begin{equation}\label{eq:secularEq2}
\begin{split}
\text{tr}\left( \D^2 \left(\D^2 + \gamma \I_p \right)^{-2}\mathbf{d}\mathbf{d}^{\rm T} \right) & \text{tr}\left( \left(\D_{1}^2 + \gamma \I_{p_1} \right)^{-2}\left(\frac{p}{p_1}\D_{1}^2 + \gamma \I_{p_1}\right) \right) + \frac{(p-p_1)}{\gamma}\text{tr}\left( \D^2 \left(\D^2 + \gamma \I_p \right)^{-2}\mathbf{d}\mathbf{d}^{\rm T} \right) \\
&-\text{tr}\left(\left(\D^2 + \gamma \I_p \right)^{-2} \mathbf{d}\mathbf{d}^{\rm T}\right)\text{tr}\left( \D_{1}^2 \left(\D_{1}^2 + \gamma \I_{p_1} \right)^{-2} \left(\frac{p}{p_1}\D_{1}^2 + \gamma \I_{p_1}\right) \right) = 0
\end{split}
\end{equation}
\hrule
\endgroup
\vspace{-2pt}
\end{figure*}

\subsection{Bounded Perturbation Regularization (BPR)}
\label{subsec:bpr}
Similar to COPRA, the BPR approach is also based on the model \eqref{eq:yDel} \cite{bpr}. The derivation of the BPR algorithm takes similar steps to those of COPRA except for the eigenvalue matrix partitioning step \eqref{eq:EVD2}, which is omitted. In fact, the BPR algorithm can be obtained by setting $p_1 = p$ and manipulating \eqref{eq:secularEq2}, which results in
\begin{align}
\label{eq:bpr}
&\text{tr} \left(\left(\D^2 + \gamma \I_p \right)^{-1}\right) \text{tr} \left(\left(\D^2 + \gamma \I_p \right)^{-1}\mathbf{d}\mathbf{d}^{\rm T}\right) \nonumber \\
& -p\,\text{tr} \left(\left(\D^2 + \gamma \I_p \right)^{-2}\mathbf{d}\mathbf{d}^{\rm T}\right)=0.
\end{align}

The above nonlinear equation can be solved using Newton's method to obtain the regularization parameter pertaining to the BPR algorithm. \\

\subsection{The Generalized Cross-validation (GCV) Method}
\label{subsec:bpr}
One may consider using the GCV for automating the regularization parameter selection for R2LDA. In contrast to COPRA and BPR, GCV hinges on a different philosophy and is based on minimizing the GCV function \cite{book_gcv}:
\begin{equation}
\label{eq:gcv}
G(\gamma)=\frac{\Big\|\Sighat^{\frac{1}{2}} (\Sighat+\gamma\I_p)^{-1}\Sighat^{\frac{1}{2}}\y -\mathbf{y}\Big\|_2^2}{\left(\text{tr} \left(\I_p-\Sighat^{\frac{1}{2}}(\Sighat+\gamma\I_p)^{-1}\Sighat^{\frac{1}{2}}\right)\right)^2},
\end{equation}
\textcolor{black}{which can be manipulated to the form}
\begin{equation}
\label{eq:gcv reduced}
\textcolor{black}{
G(\gamma)=\frac{\Big\|\D^2(\D^2+\gamma\I_p)^{-1}\mathbf{d} -\mathbf{d} \Big\|_2^2}{\left(p - \text{tr} \left(\D^2(\D^2+\gamma\I_p)^{-1}\right)\right)^2}.
}
\end{equation}

The GCV approach can be thought of as an approximation of leave-one-out cross-validation (the reader can refer to \cite{book_gcv}, chapter 4). To compute the regularization parameter using the GCV, a line search that evaluates $G(\gamma)$ over a suitably chosen $\gamma$ interval is carried out. To set up the interval, we apply the technique described in \cite{Hansen2007}.

\subsection{Summary of the Proposed R2LDA Approach}
\label{subsec:rrlda_algo}
The main steps involved in the proposed R2LDA approach are summarized as follows:
\begin{enumerate}[font={\bfseries}, label={\arabic*)}]
    \item {Estimate the class statistics $\Muhat_i$, $\Sighat_i$ and $\Sighat$ from the training data by using (\ref{eq:MuSig_est}) and (\ref{eq:Sighat}).}
    \item  {Compute $\Muhat^+$, $\Muhat^-$ and the EVD of $\Sighat$.}
    \item {Set $\y\!=\!\Muhat^-$ in the model (\ref{eq:ySig}) and obtain $\gamma_b$ using the chosen regularization parameter selection method.}
    \item {For a given test sample, compute $\x'$.}
    \item {Set $\y \!=\! \x'$ in the model (\ref{eq:ySig}) and obtain $\gamma_z$ using the chosen regularization parameter selection method.}
    \item {Compute the R2LDA score function using (\ref{eq:RRLDAhat3}), and assign the test sample to a class according to (\ref{eq:LDArule}).}
\end{enumerate}

\textcolor{black}{In Step~3 and Step~5, we apply any of the three regularization parameter selection methods discussed in the previous subsections (COPRA, BPR or GCV).} Henceforth, the resulting classification algorithm will be referred to as COPRA-R2LDA, BPR-R2LDA, or GCV-R2LDA, \textcolor{black}{depending on the regularization parameter selection method used}.

\section{Performance Evaluation}
\label{sec:sim}
We demonstrate the performance of the proposed R2LDA classifiers with different regularization parameter selection techniques against the RLDA classifiers of the asymptotic error estimator (Asym-RLDA)\cite{zollanvari2} and the optimal-intercept-improved RLDA (OII-RLDA) \cite{Sifaou2020}. We consider both synthetic and real data for performance evaluation. The codes used to generate the results are available online\footnote{\url{https://kaust-my.sharepoint.com/:f:/g/personal/ahmedt_kaust_edu_sa/EpVhsbg3Dw9IgIJ1KT7sSxkB_5tJSPMMJ0lDADAndl-sTQ?e=bWdIRT}}.

\textcolor{black}{We use the average percentage classification error as the performance metric. This section also discusses the computational complexity of various algorithms.}

\begin{figure*}[!t]
\begin{center}
\subfloat[Gaussian, $\sigma=0$]{\includegraphics[width=0.3\textwidth]{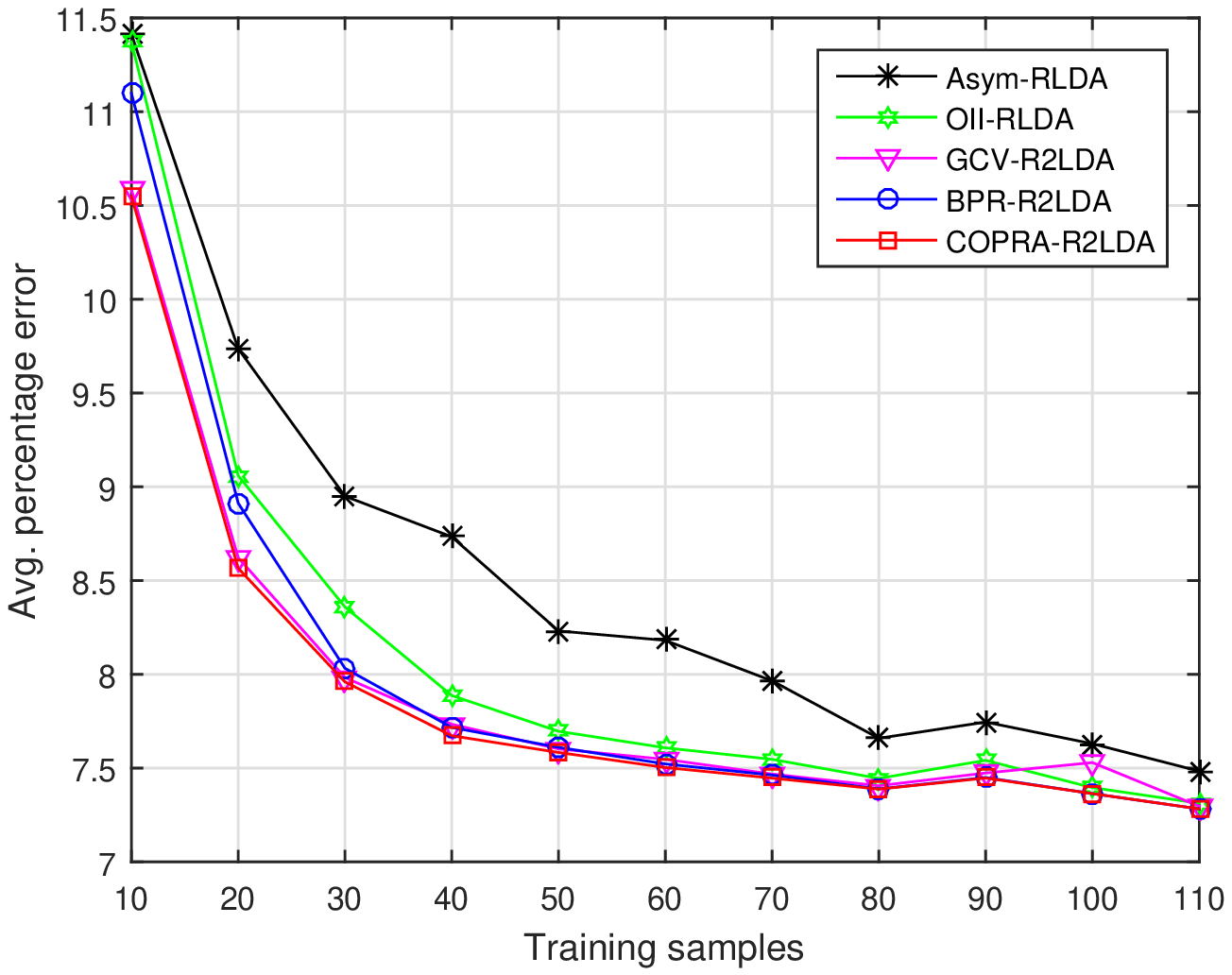}}
\subfloat[Gaussian, $\sigma=0.1$]{\includegraphics[width=0.3\textwidth]{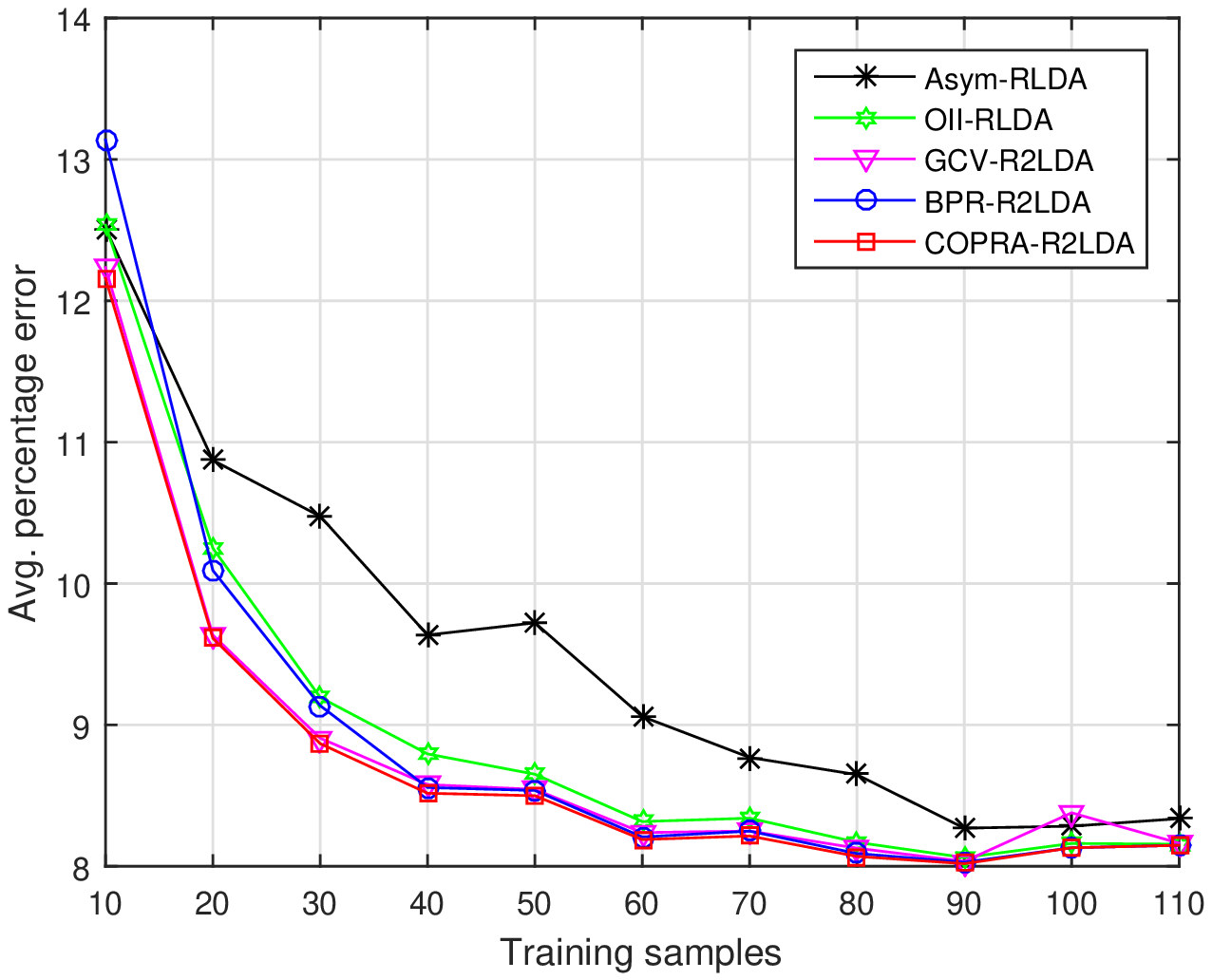}}
\subfloat[Gaussian, $\sigma=0.2$]{\includegraphics[width=0.3\textwidth]{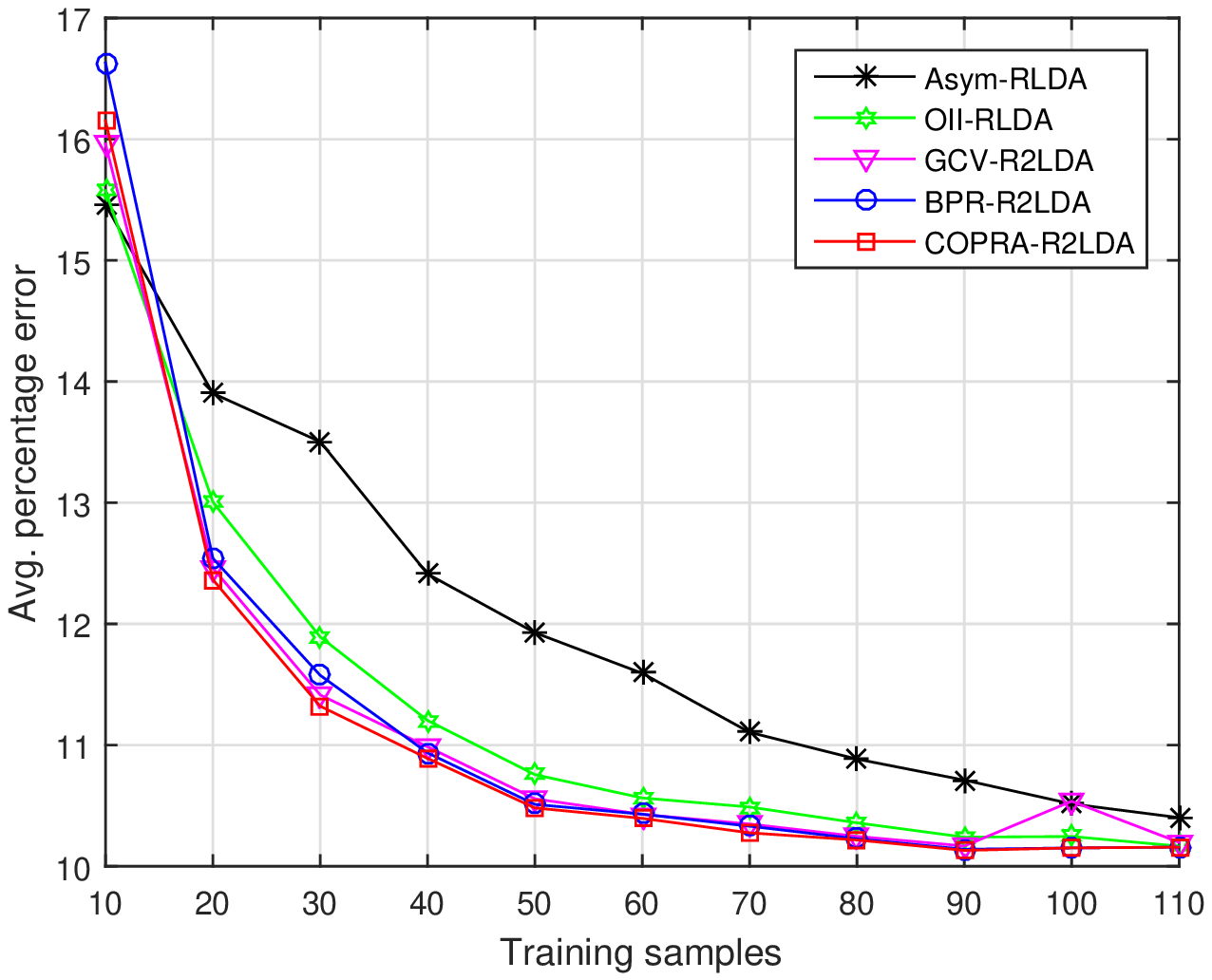}}
\caption{Gaussian data misclassification rates versus training data size for different test data noise levels.}
\label{fig:Gauss}
\end{center}
\end{figure*}

\begin{figure*}[!t]
\begin{center}
\subfloat[MNIST (1,7), $\sigma = 0$]{\includegraphics[width=0.3\textwidth]{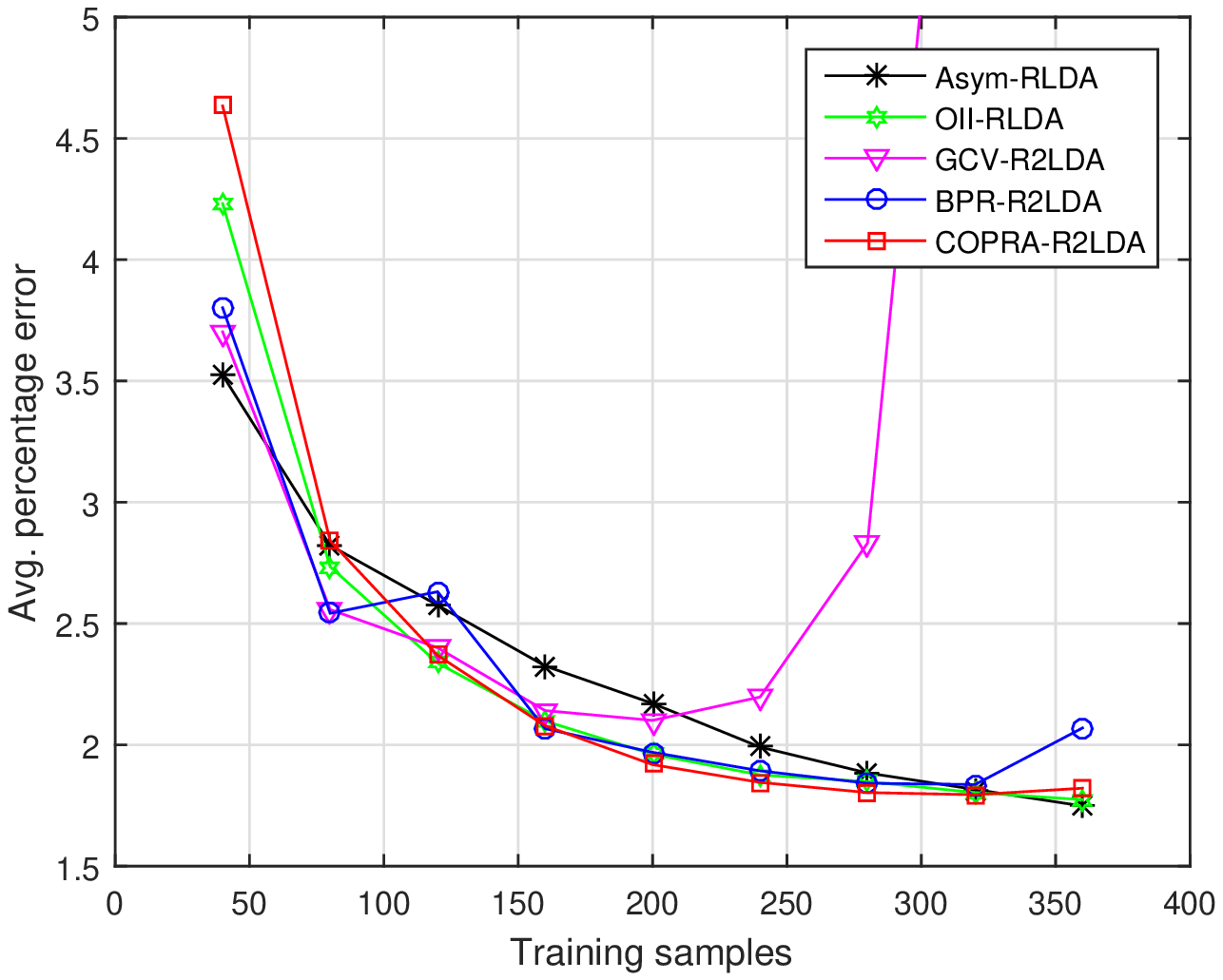}}
\subfloat[MNIST (1,7), $\sigma=1$]{\includegraphics[width=0.3\textwidth]{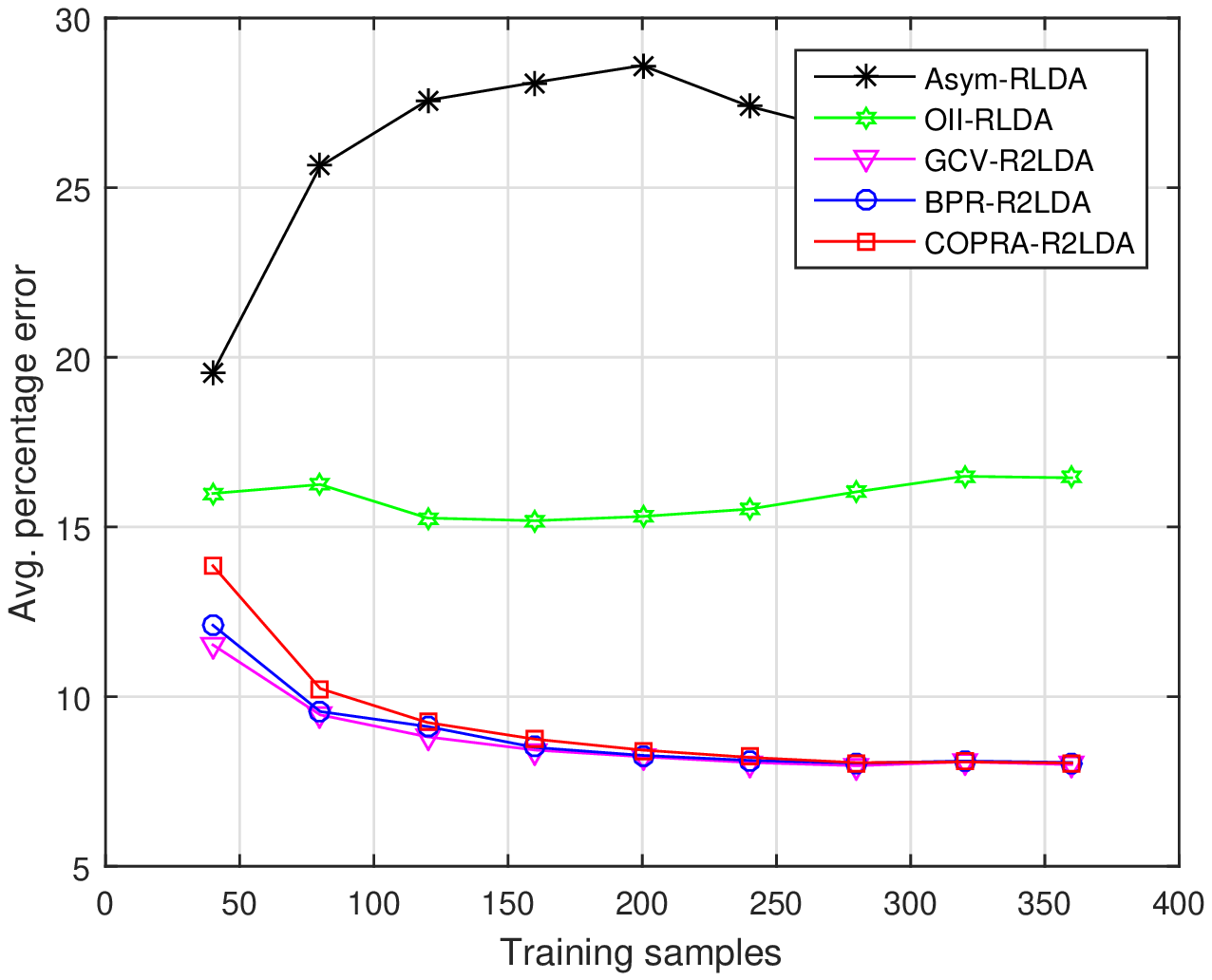}}
\subfloat[MNIST (1,7), $\sigma =2$]{\includegraphics[width=0.3\textwidth]{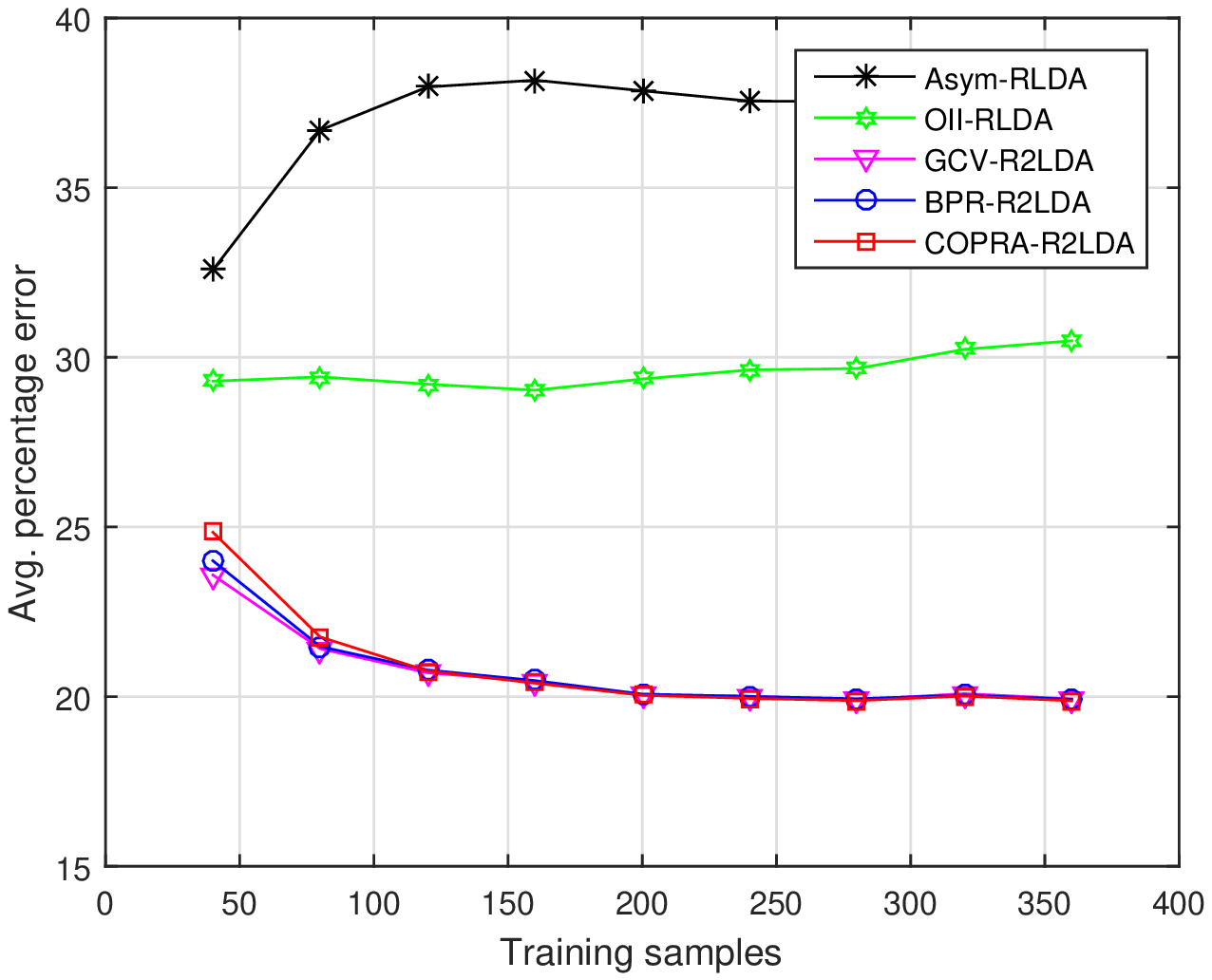}}\\
\subfloat[MNIST (5,8), $\sigma = 0$]{\includegraphics[width=0.3\textwidth]{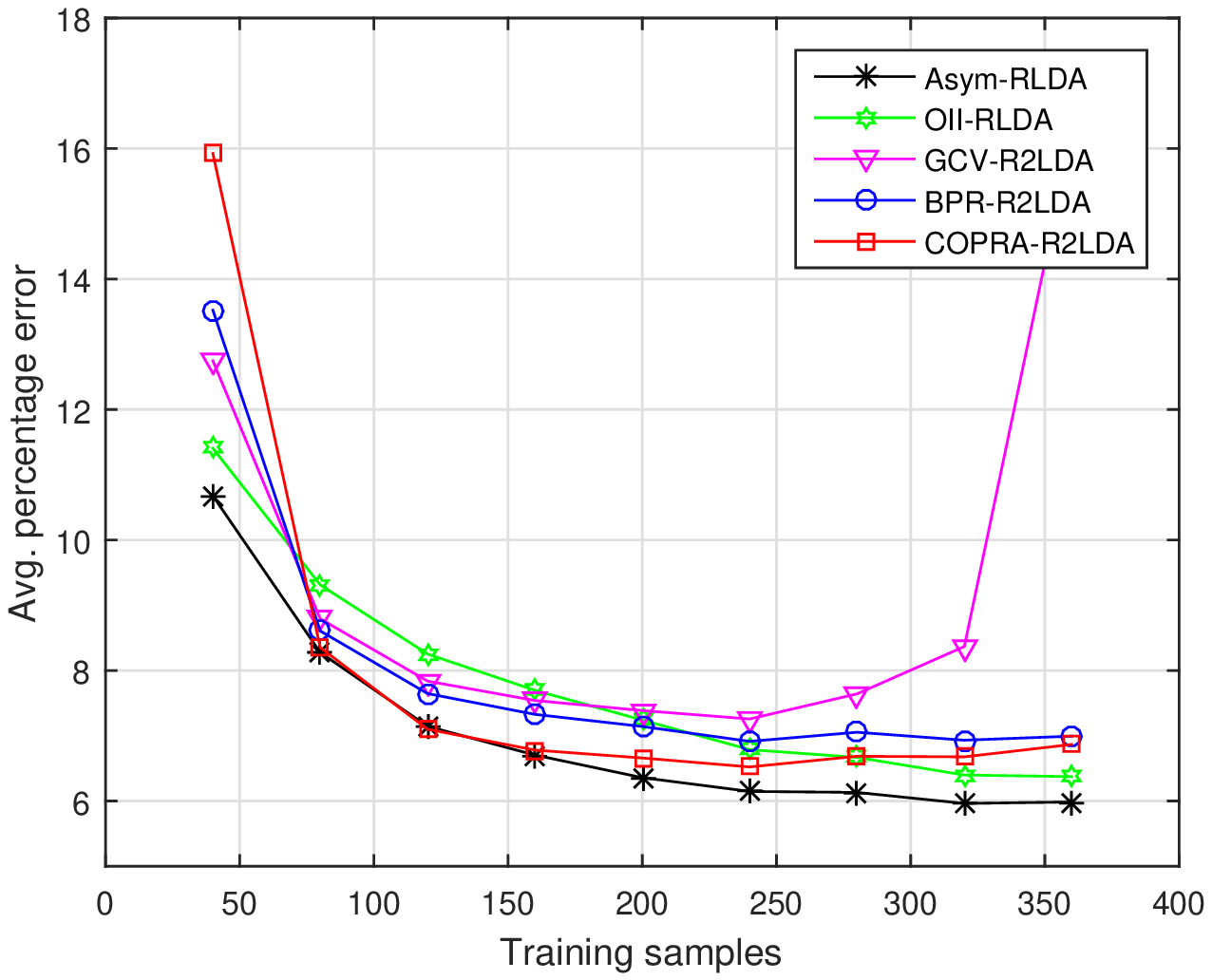}}
\subfloat[MNIST (5,8), $\sigma=1$]{\includegraphics[width=0.3\textwidth]{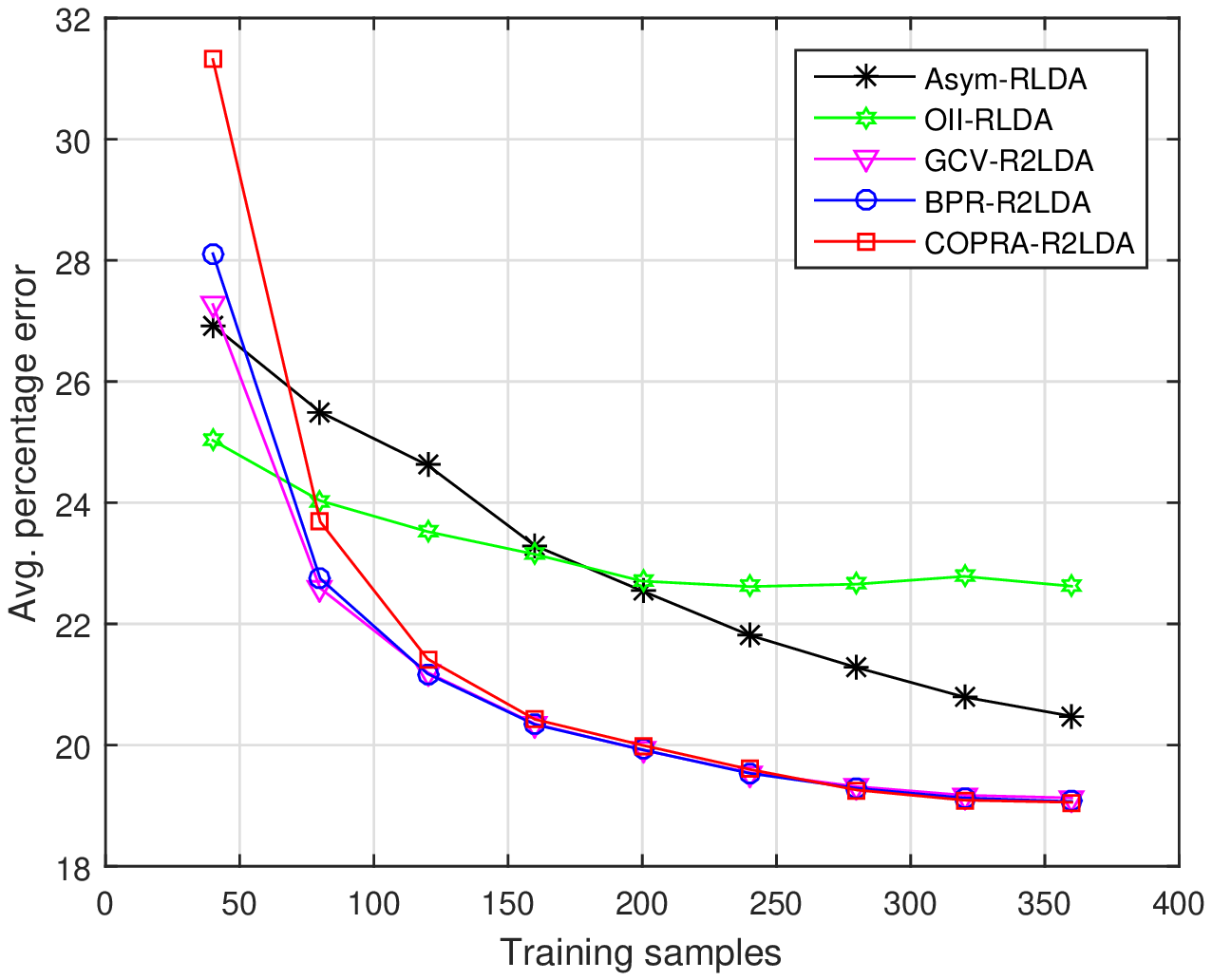}}
\subfloat[MNIST (5,8), $\sigma =2$]{\includegraphics[width=0.3\textwidth]{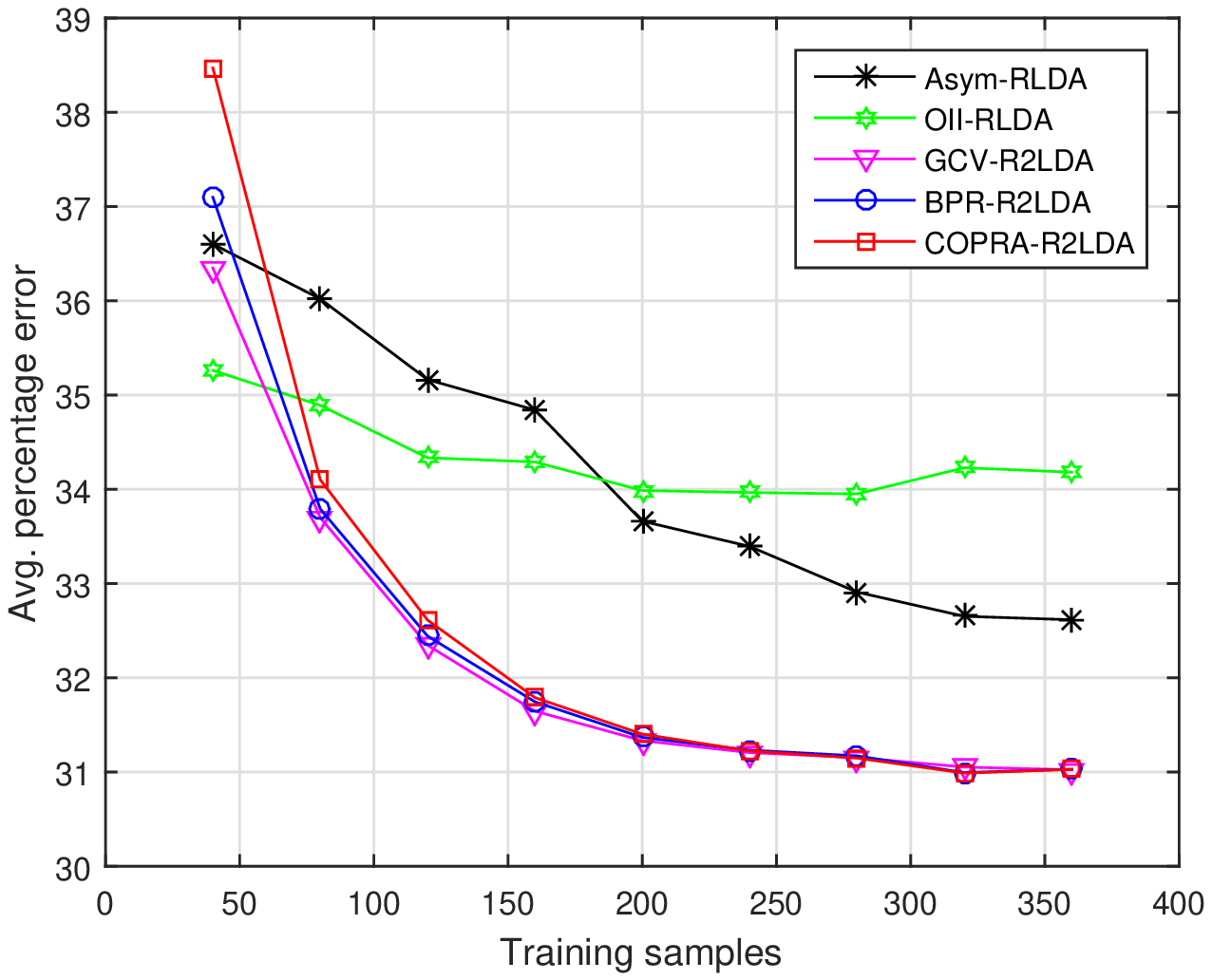}}\\
\subfloat[MNIST (7,9), $\sigma = 0$]{\includegraphics[width=0.3\textwidth]{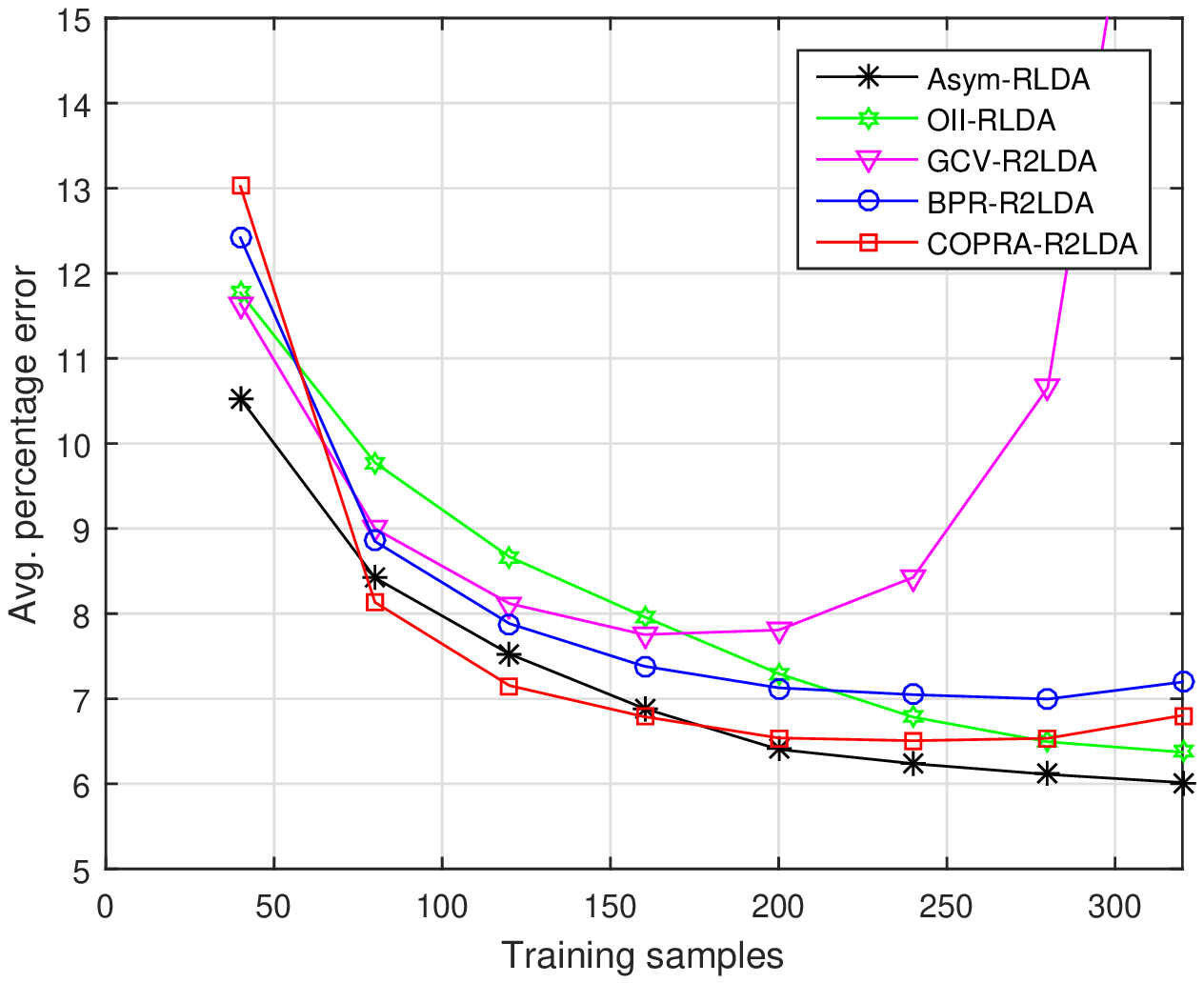}}
\subfloat[MNIST (7,9), $\sigma=1$]{\includegraphics[width=0.3\textwidth]{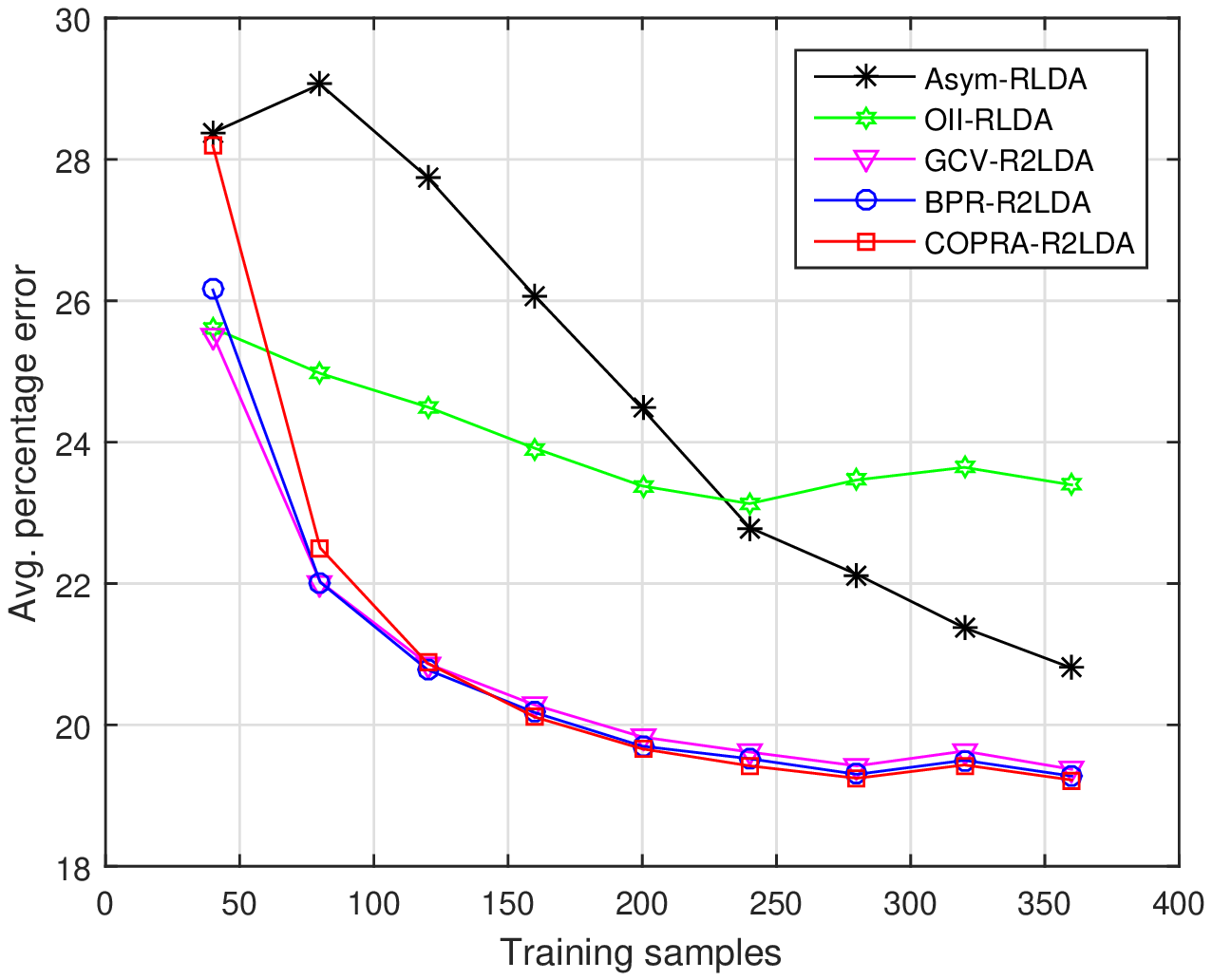}}
\subfloat[MNIST (7,9), $\sigma =2$]{\includegraphics[width=0.3\textwidth]{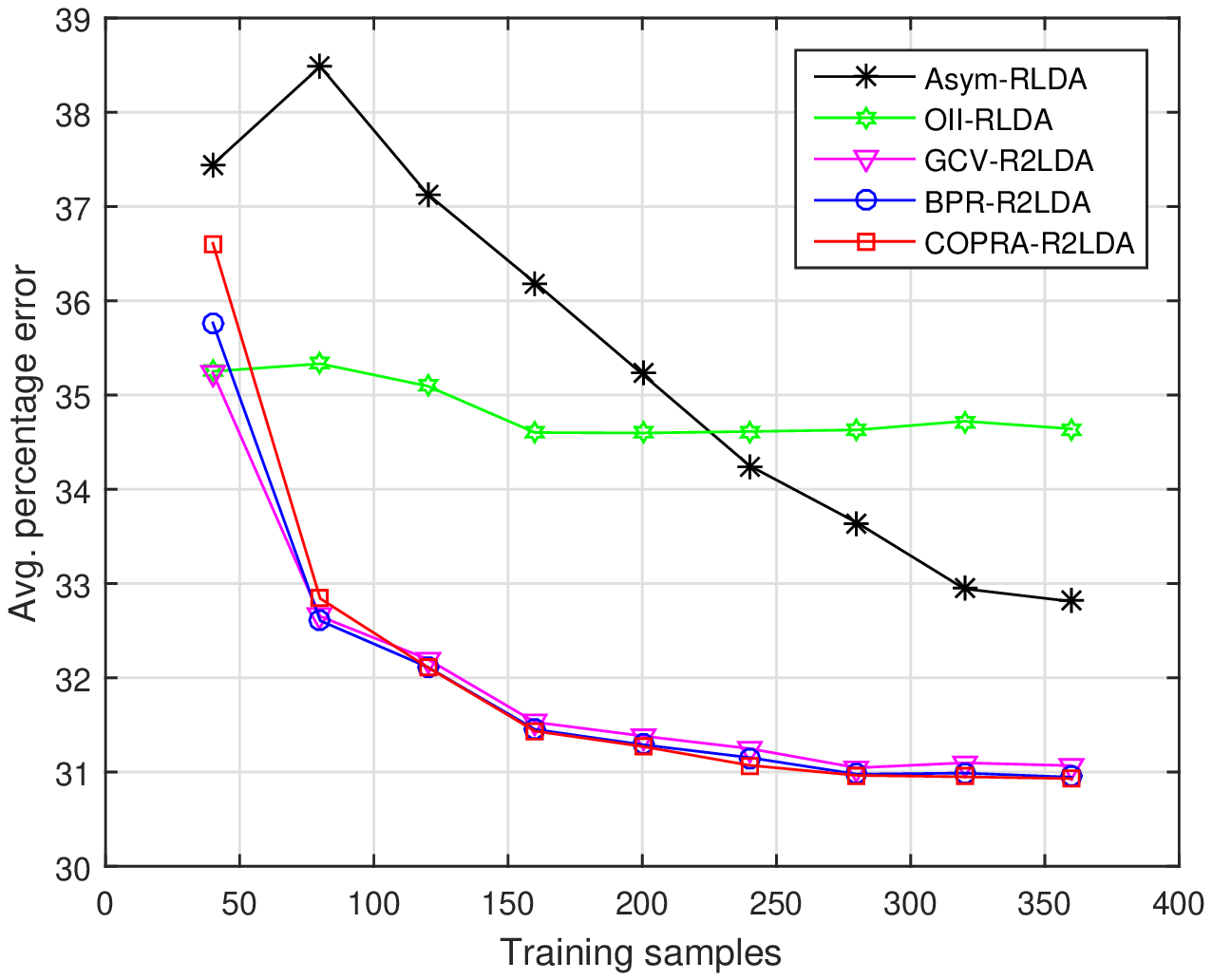}}
\caption{MNIST data misclassification rates versus training data size for different test data noise levels.}
\label{fig:MNIST}
\end{center}
\end{figure*}

\begin{figure*}[!t]
\begin{center}
\subfloat[Phonemes (1,2), $\sigma=0$]{\includegraphics[width=0.3\textwidth]{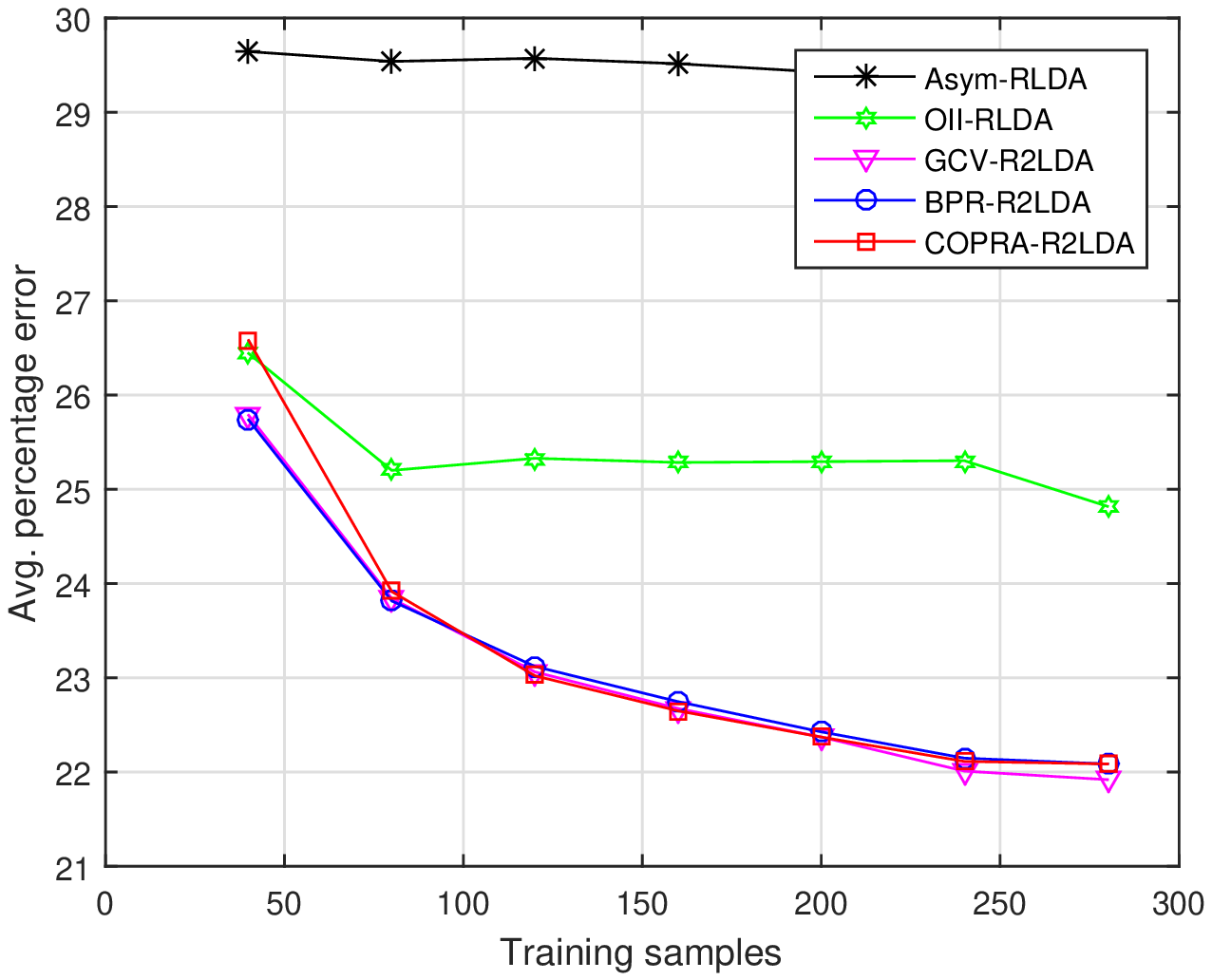}}
\subfloat[Phonemes (1,2), $\sigma=0.1$]{\includegraphics[width=0.3\textwidth]{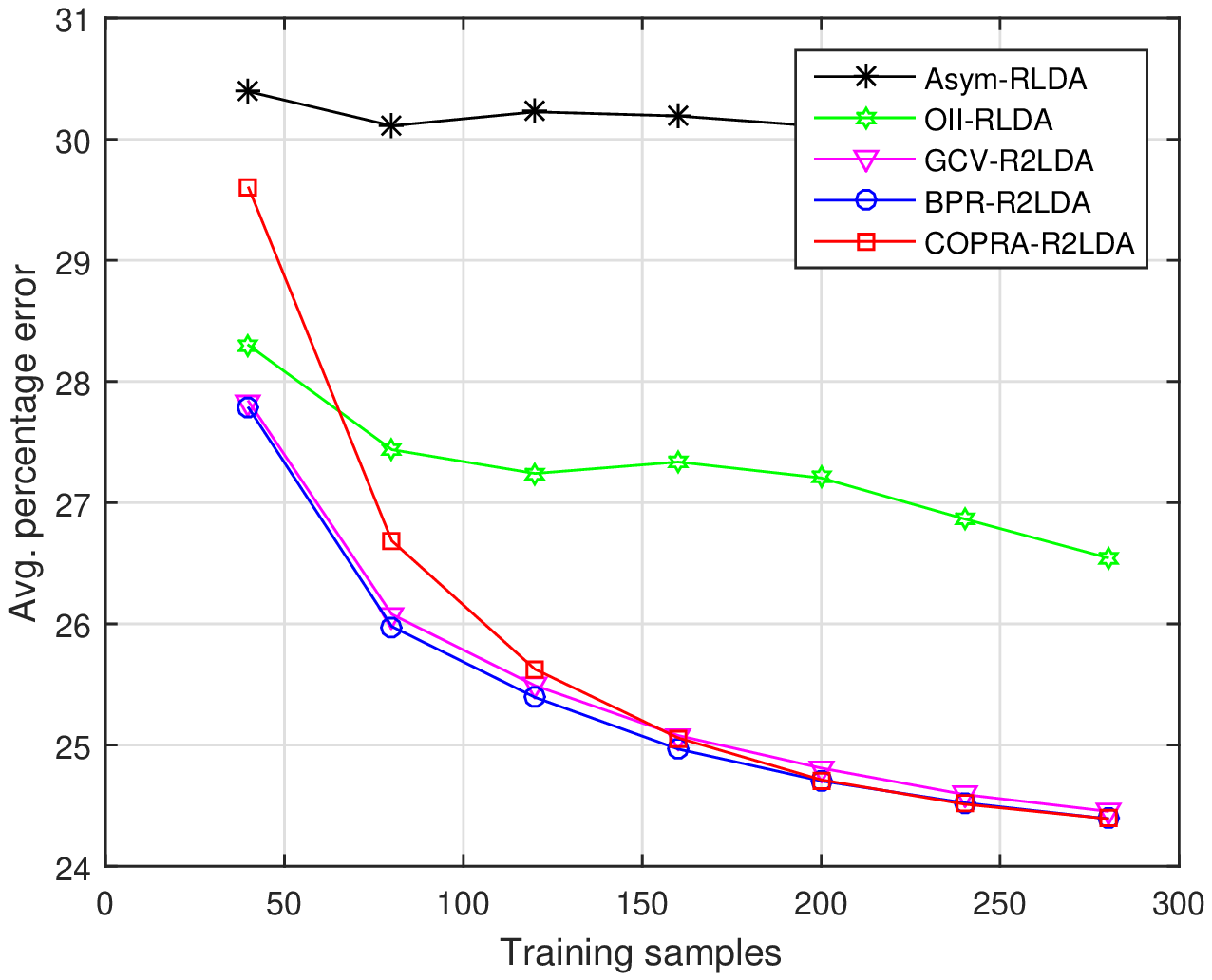}}
\subfloat[Phonemes (1,2), $\sigma=0.2$]{\includegraphics[width=0.3\textwidth]{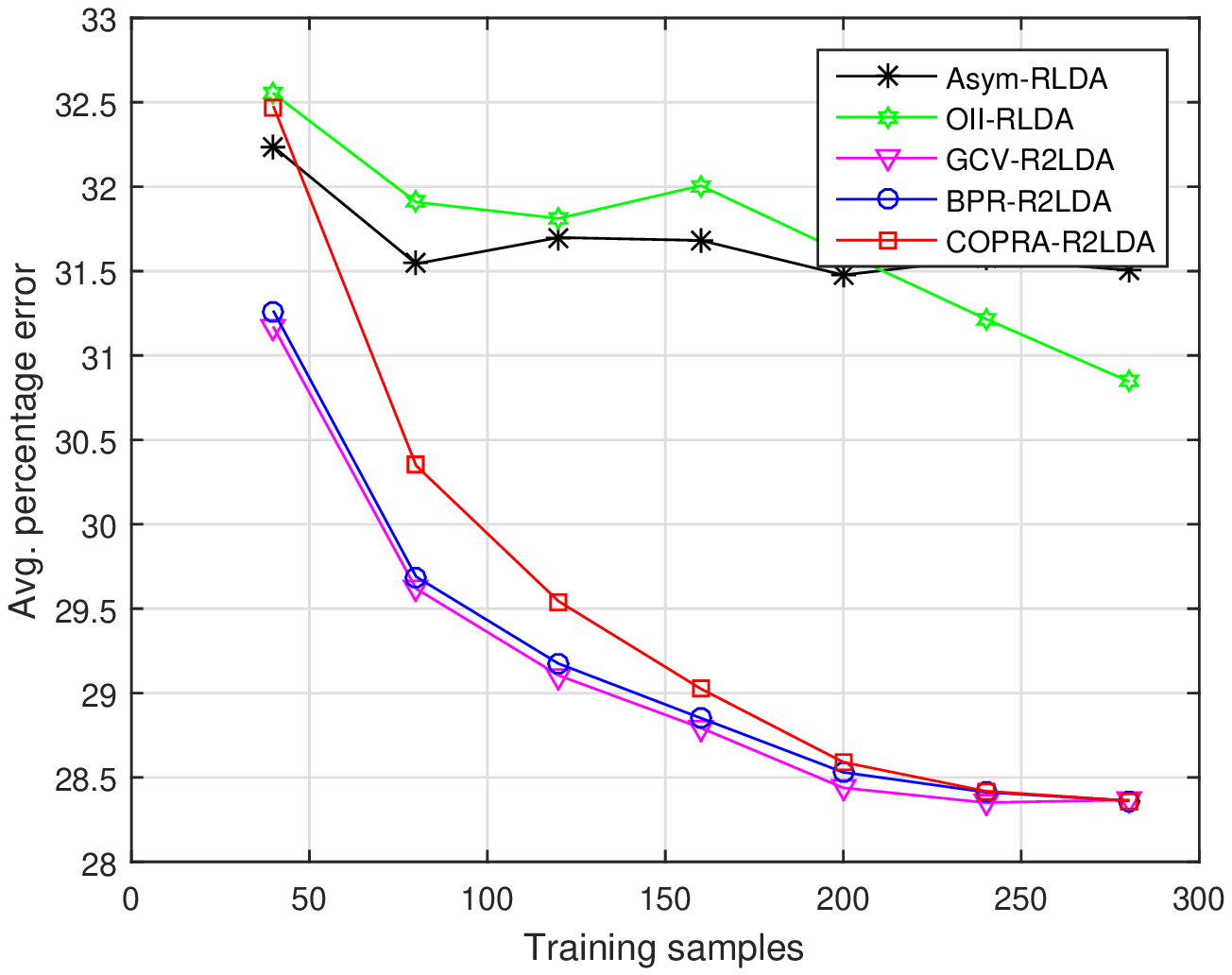}}\\
\subfloat[Phonemes (1,3), $\sigma=0$]{\includegraphics[width=0.3\textwidth]{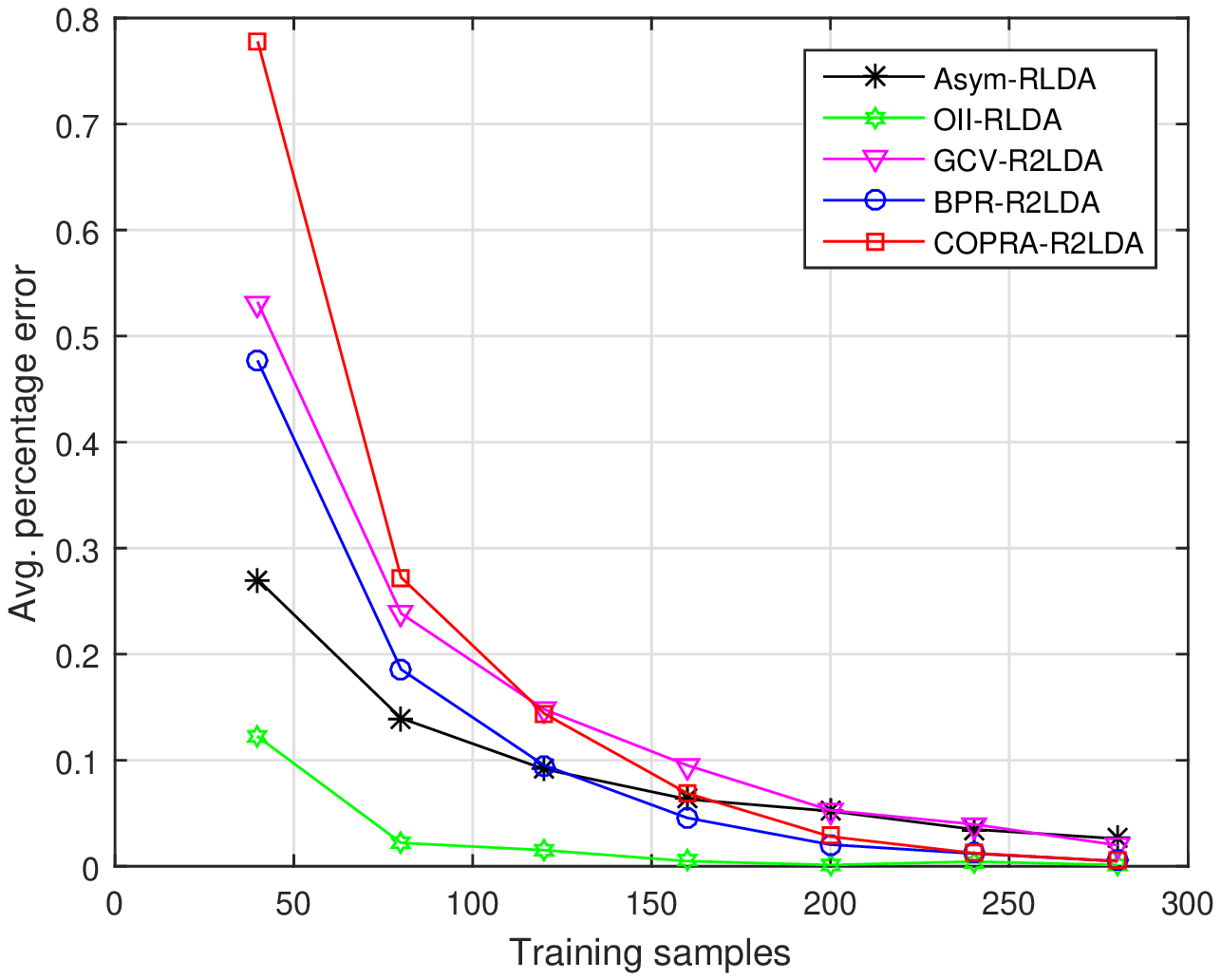}}
\subfloat[Phonemes (1,3), $\sigma=0.1$]{\includegraphics[width=0.3\textwidth]{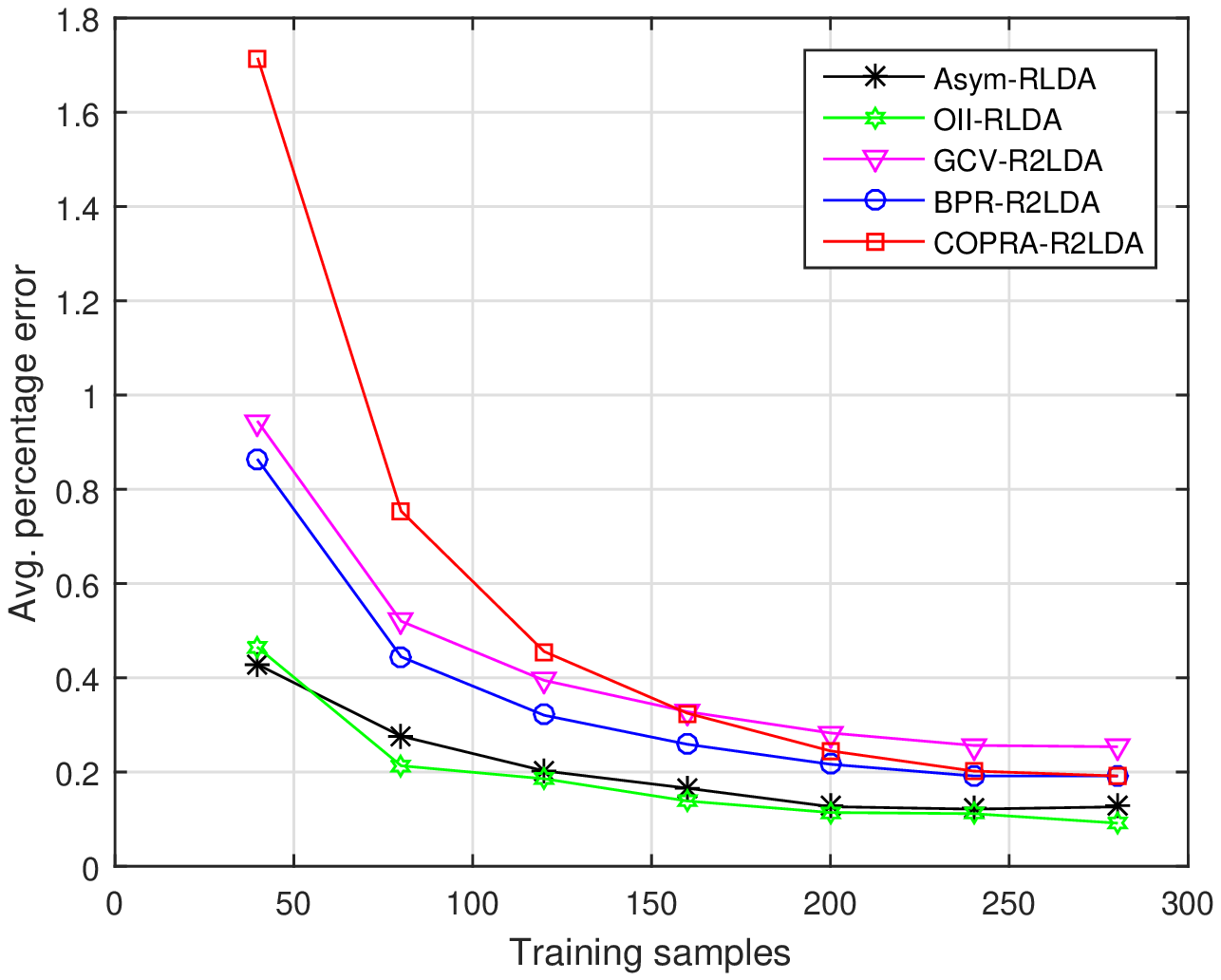}}
\subfloat[Phonemes (1,3), $\sigma=0.2$]{\includegraphics[width=0.3\textwidth]{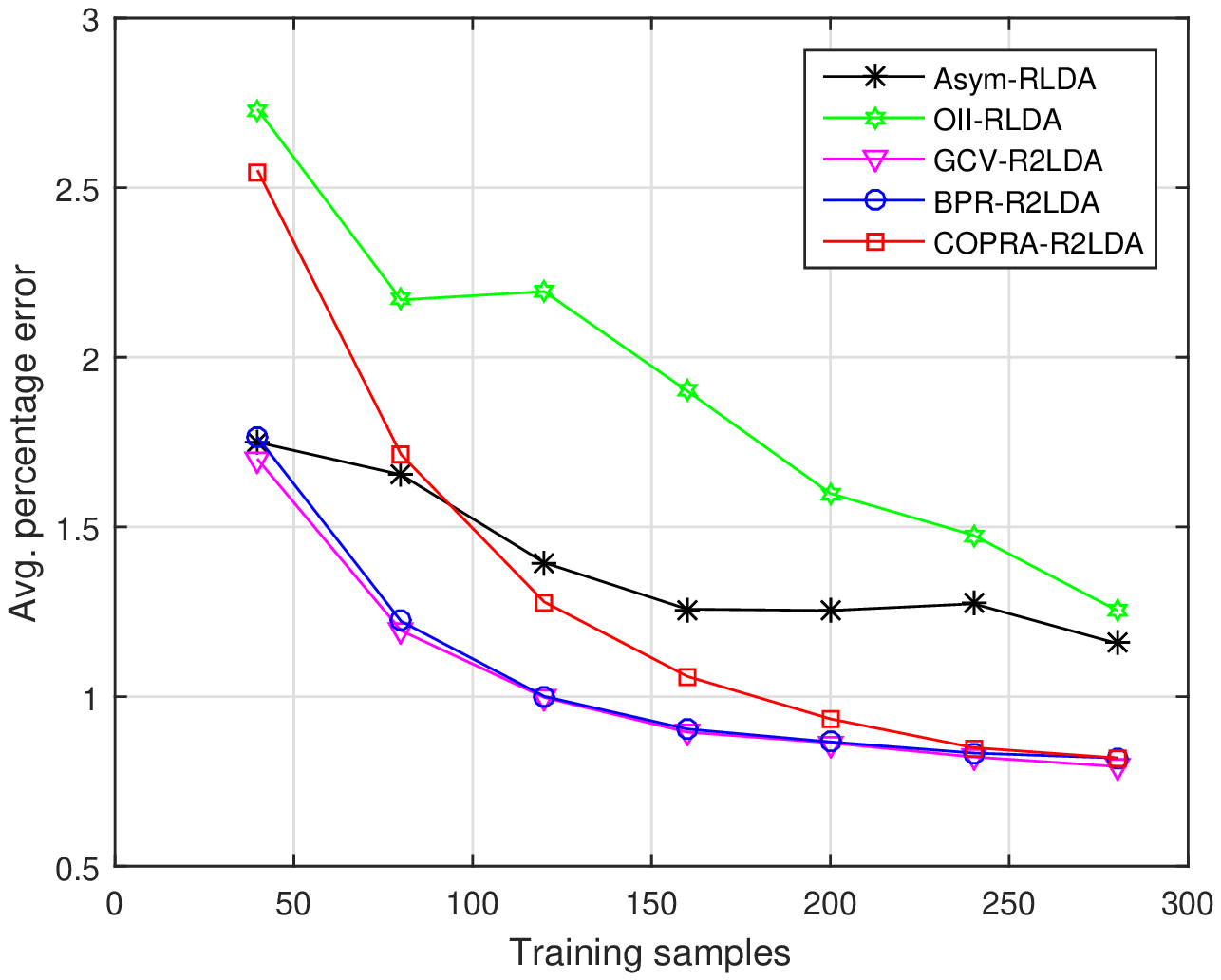}}\\
\subfloat[Phonemes (1,5), $\sigma=0$]{\includegraphics[width=0.3\textwidth]{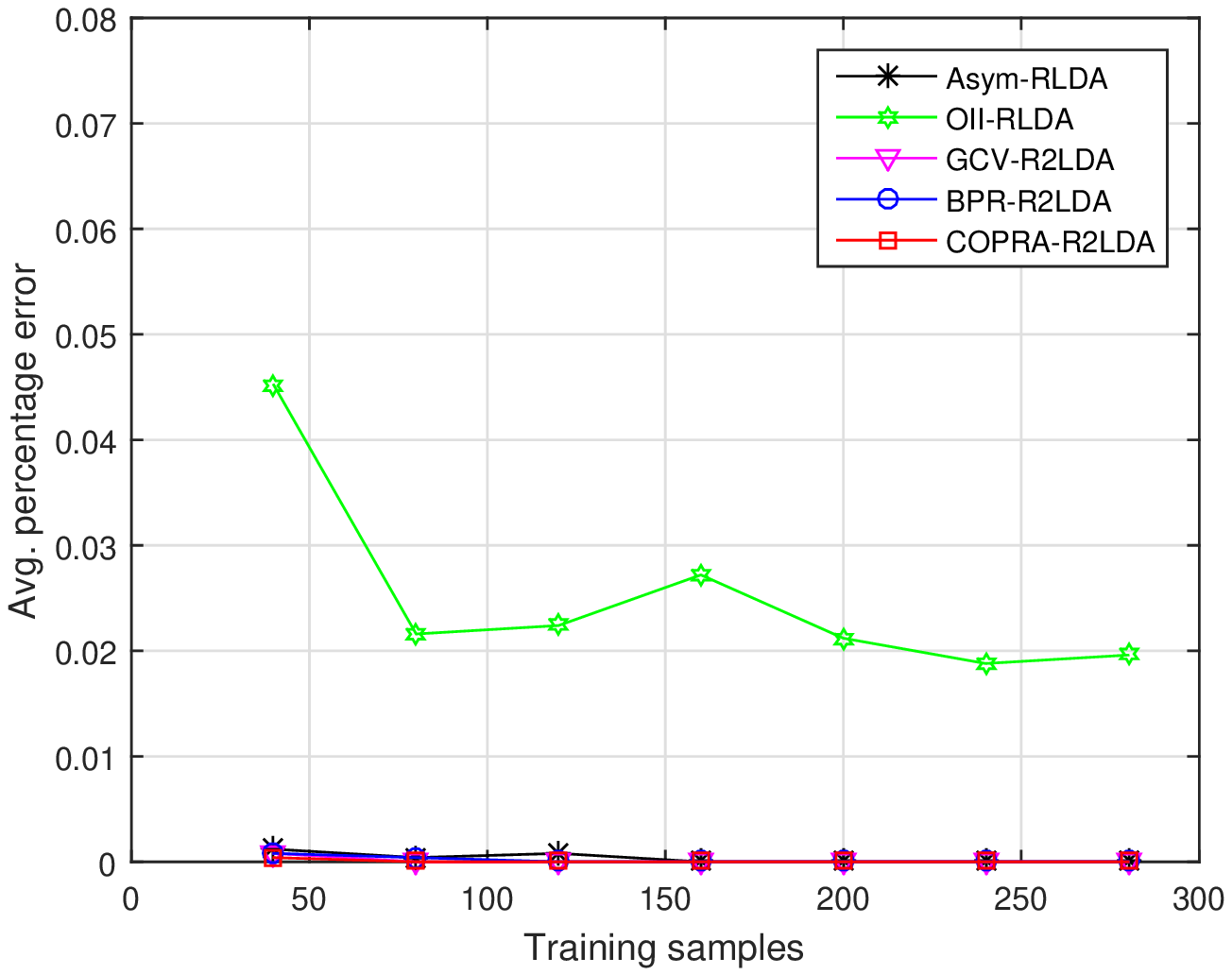}}
\subfloat[Phonemes (1,5), $\sigma=0.1$]{\includegraphics[width=0.3\textwidth]{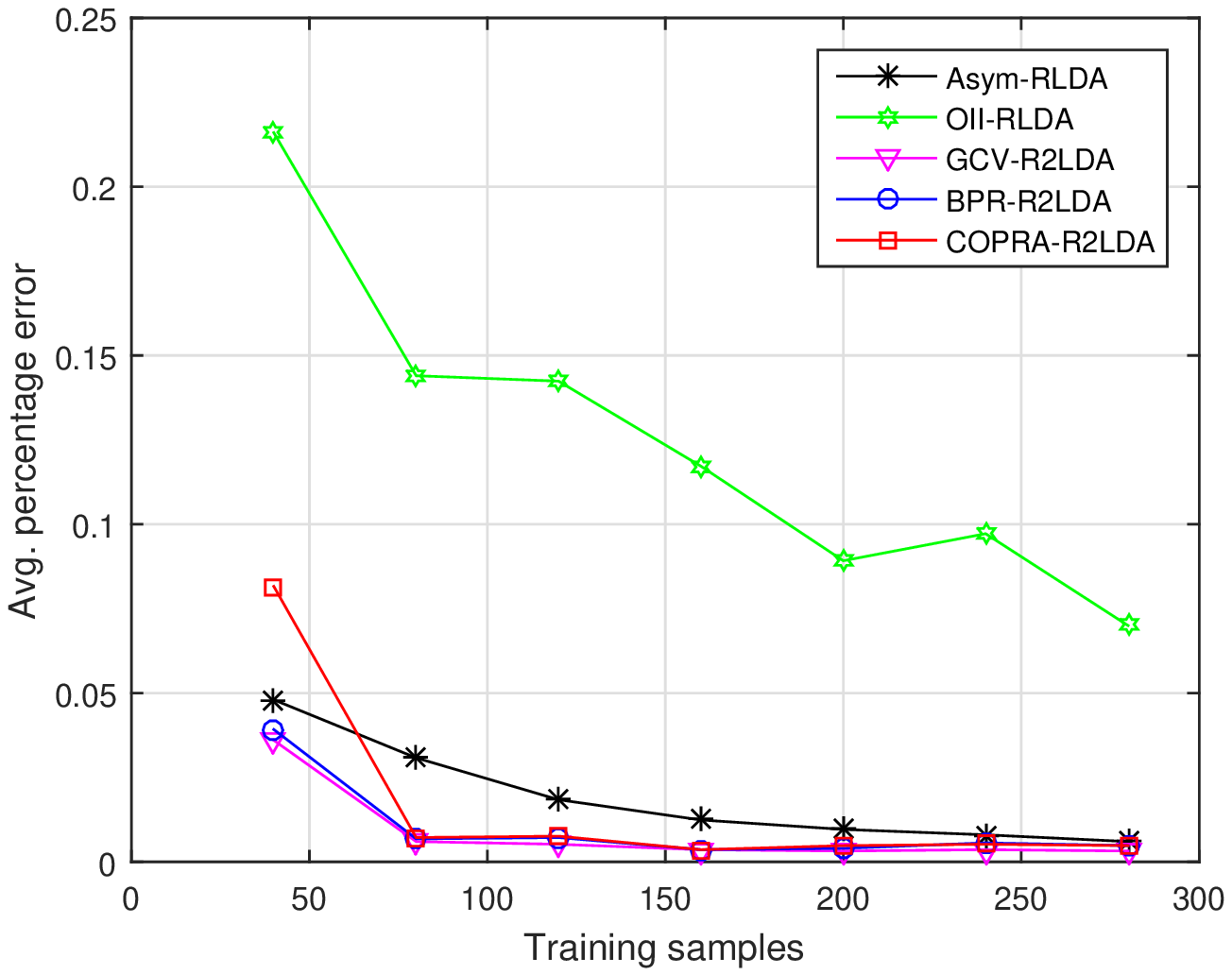}}
\subfloat[Phonemes (1,5), $\sigma=0.2$]{\includegraphics[width=0.3\textwidth]{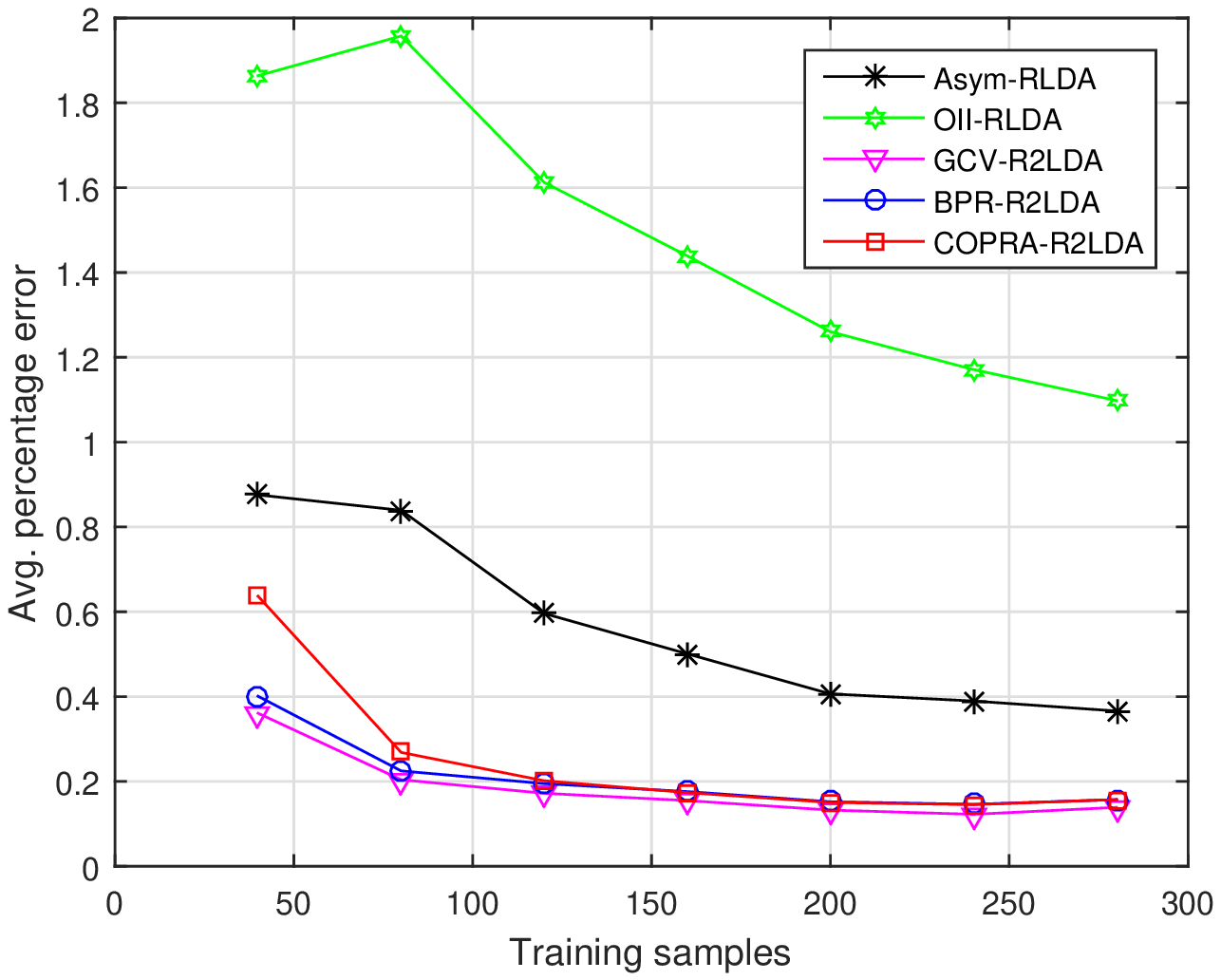}}\\
\subfloat[Phonemes (4,5), $\sigma=0$]{\includegraphics[width=0.3\textwidth]{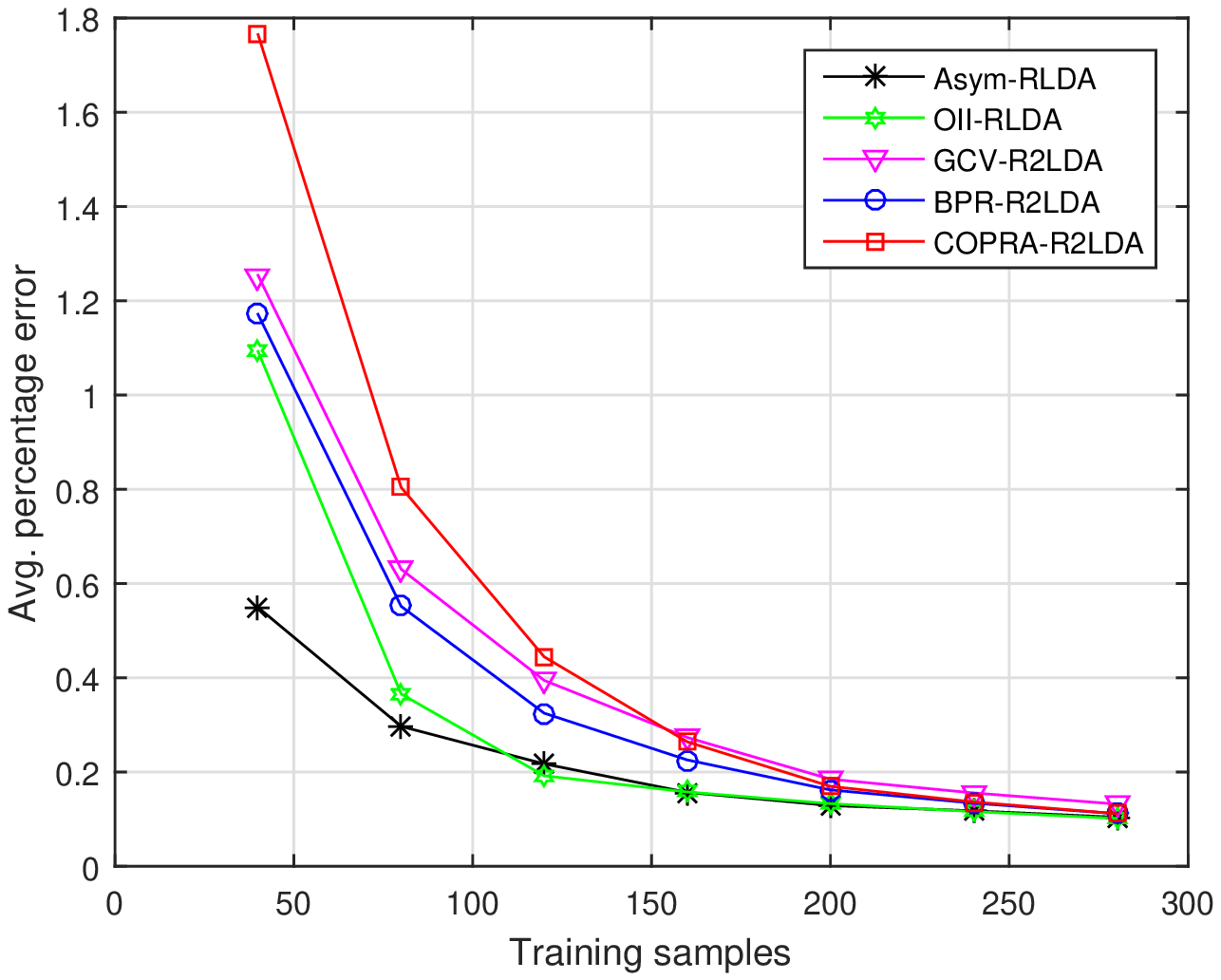}}
\subfloat[Phonemes (4,5), $\sigma=0.1$]{\includegraphics[width=0.3\textwidth]{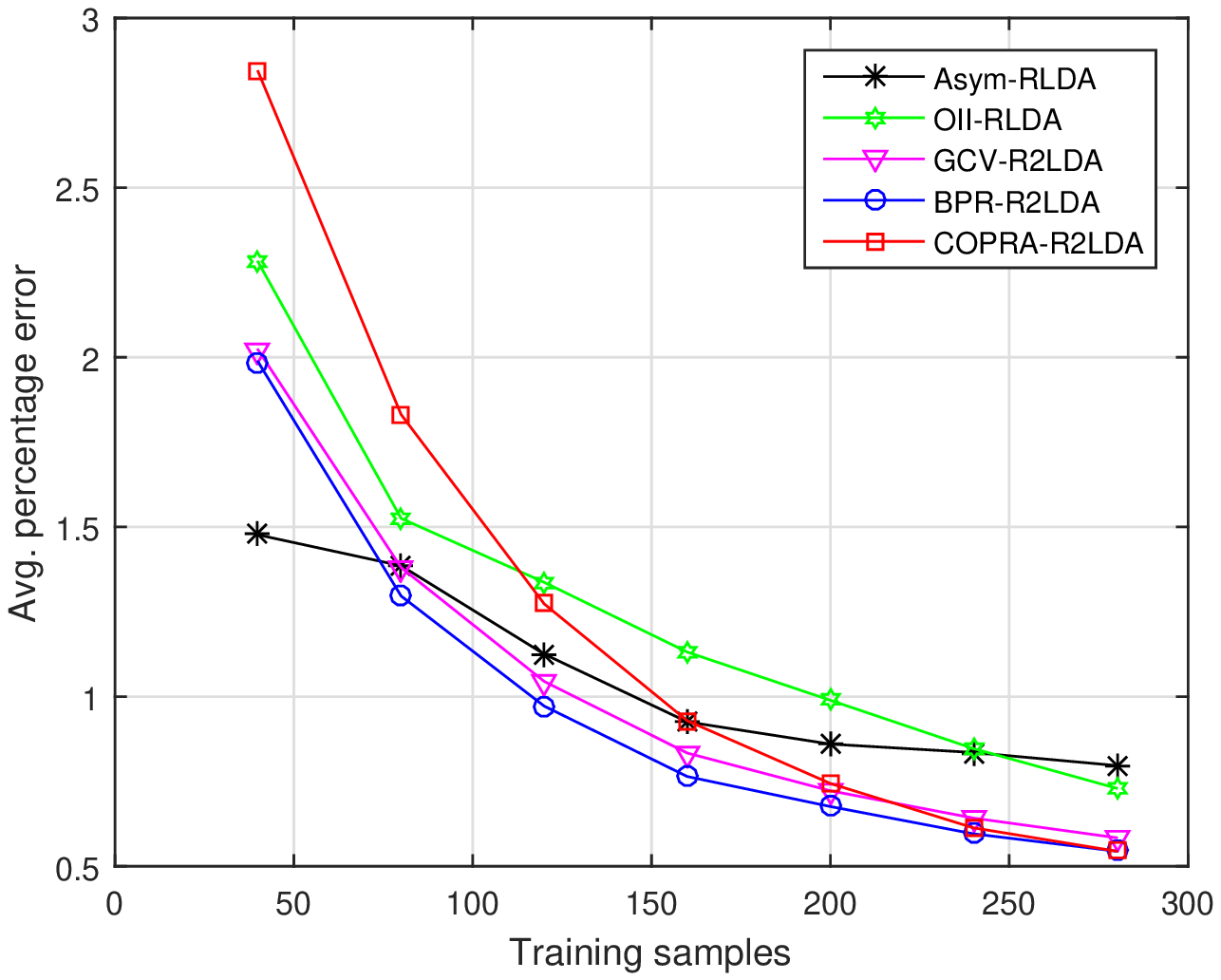}}
\subfloat[Phonemes (4,5), $\sigma=0.2$]{\includegraphics[width=0.3\textwidth]{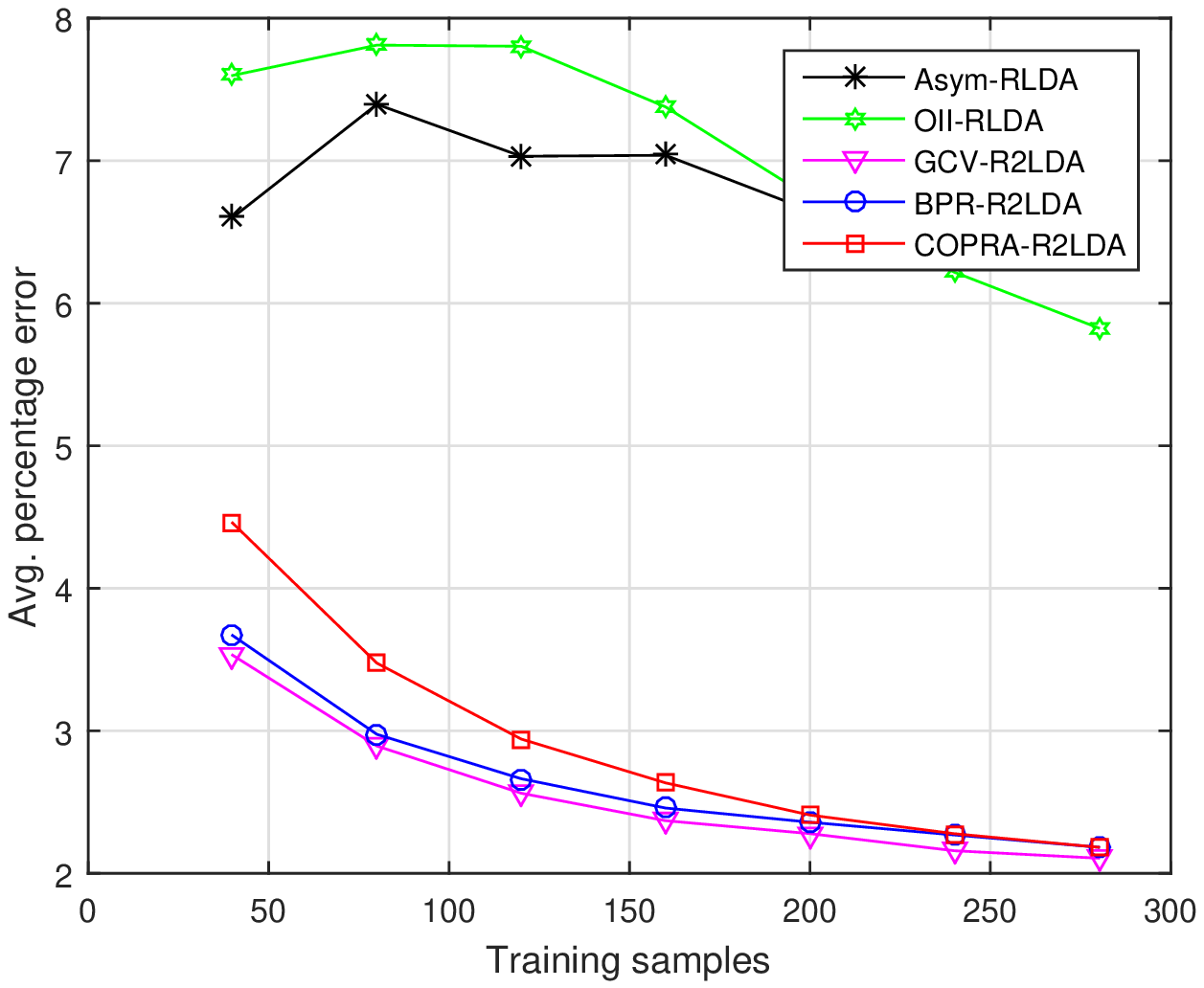}}\\
\caption{Phonemes data misclassification rates versus training data size for different test data noise levels.}
\label{fig:Phoneme}
\end{center}
\end{figure*}

\begin{figure*}[!t]
\begin{center}
\subfloat[Sonar, $\sigma=0$]{\includegraphics[width=0.3\textwidth]{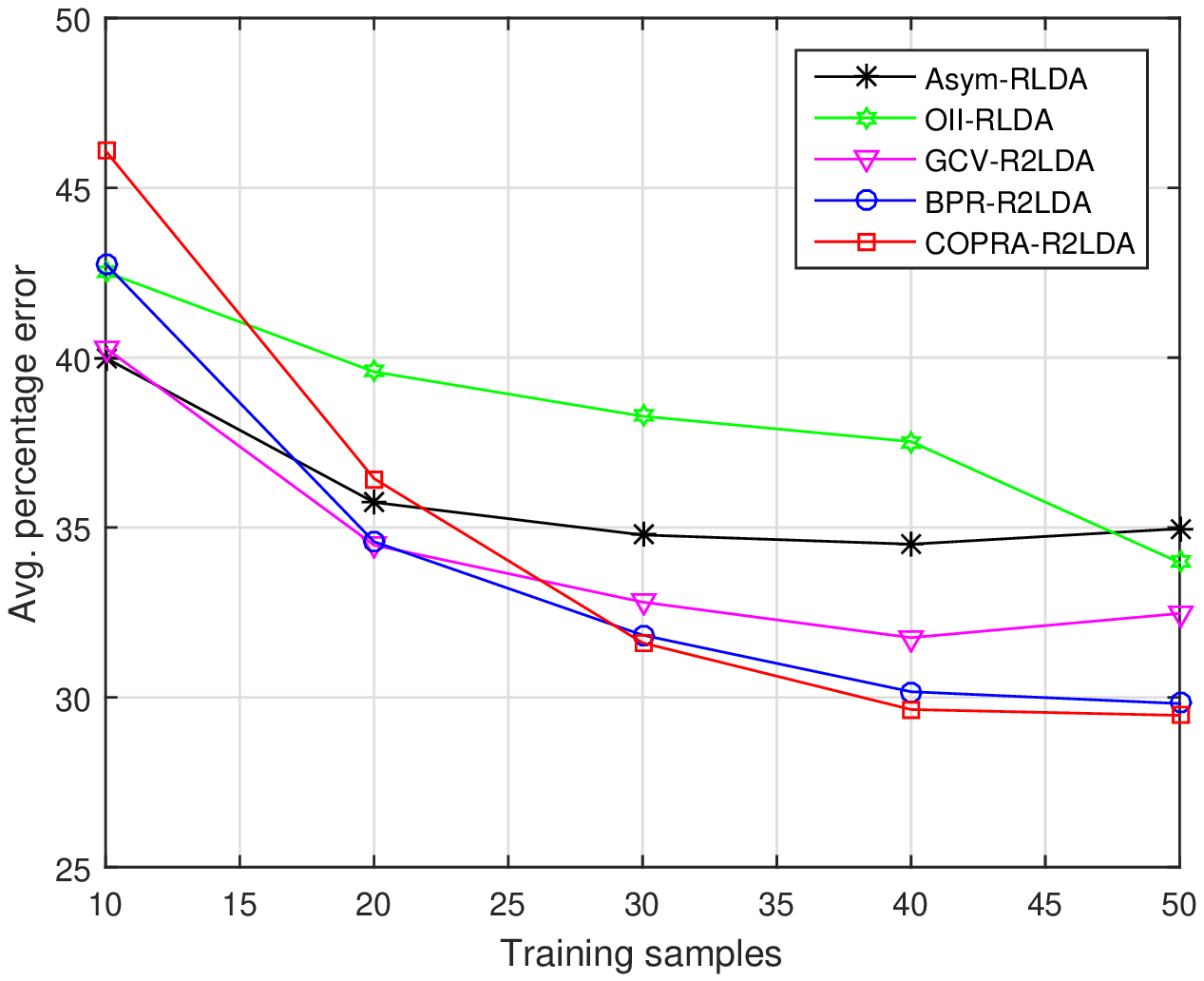}}
\subfloat[Sonar, $\sigma=0.1$]{\includegraphics[width=0.3\textwidth]{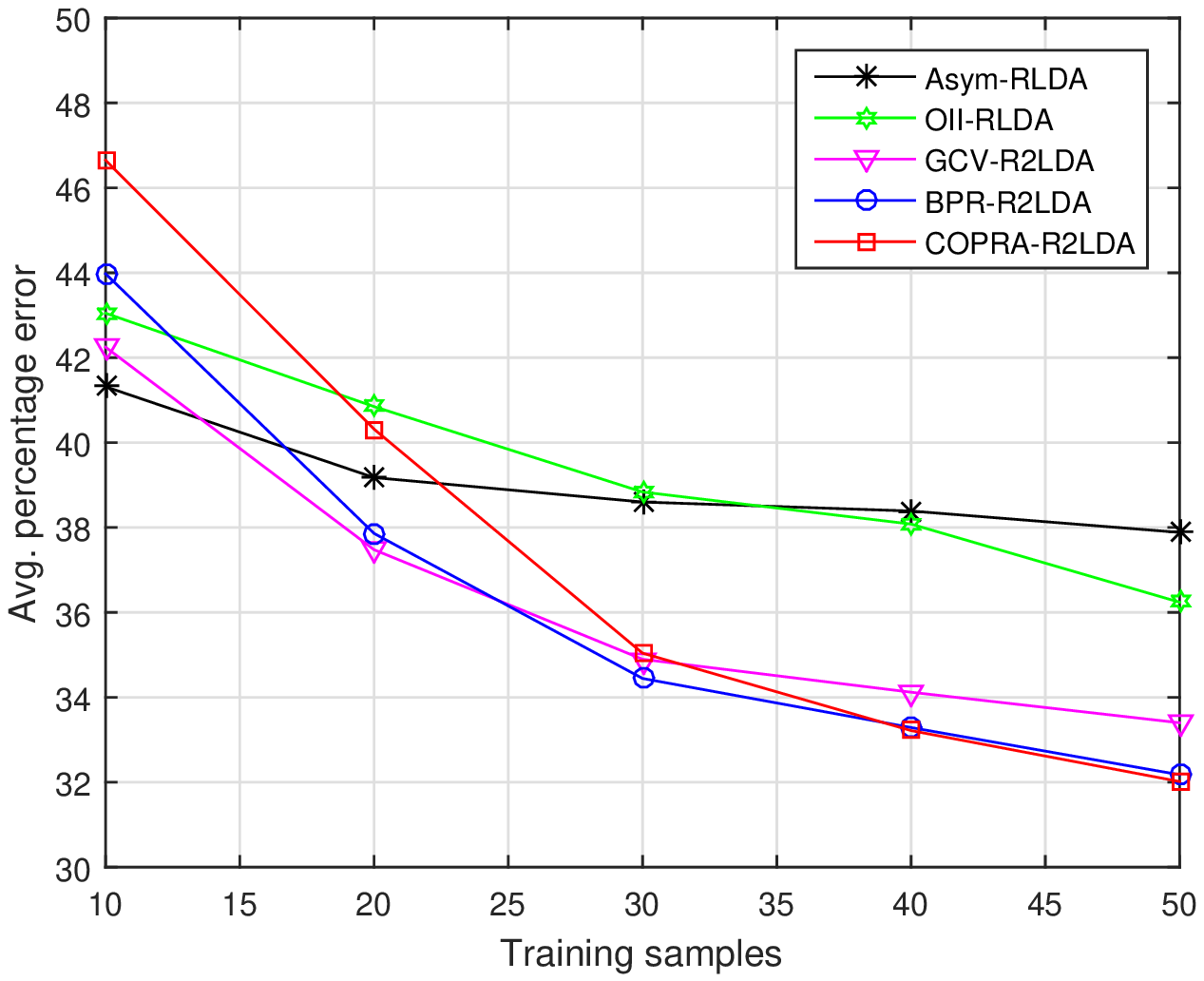}}
\subfloat[Sonar, $\sigma=0.2$]{\includegraphics[width=0.3\textwidth]{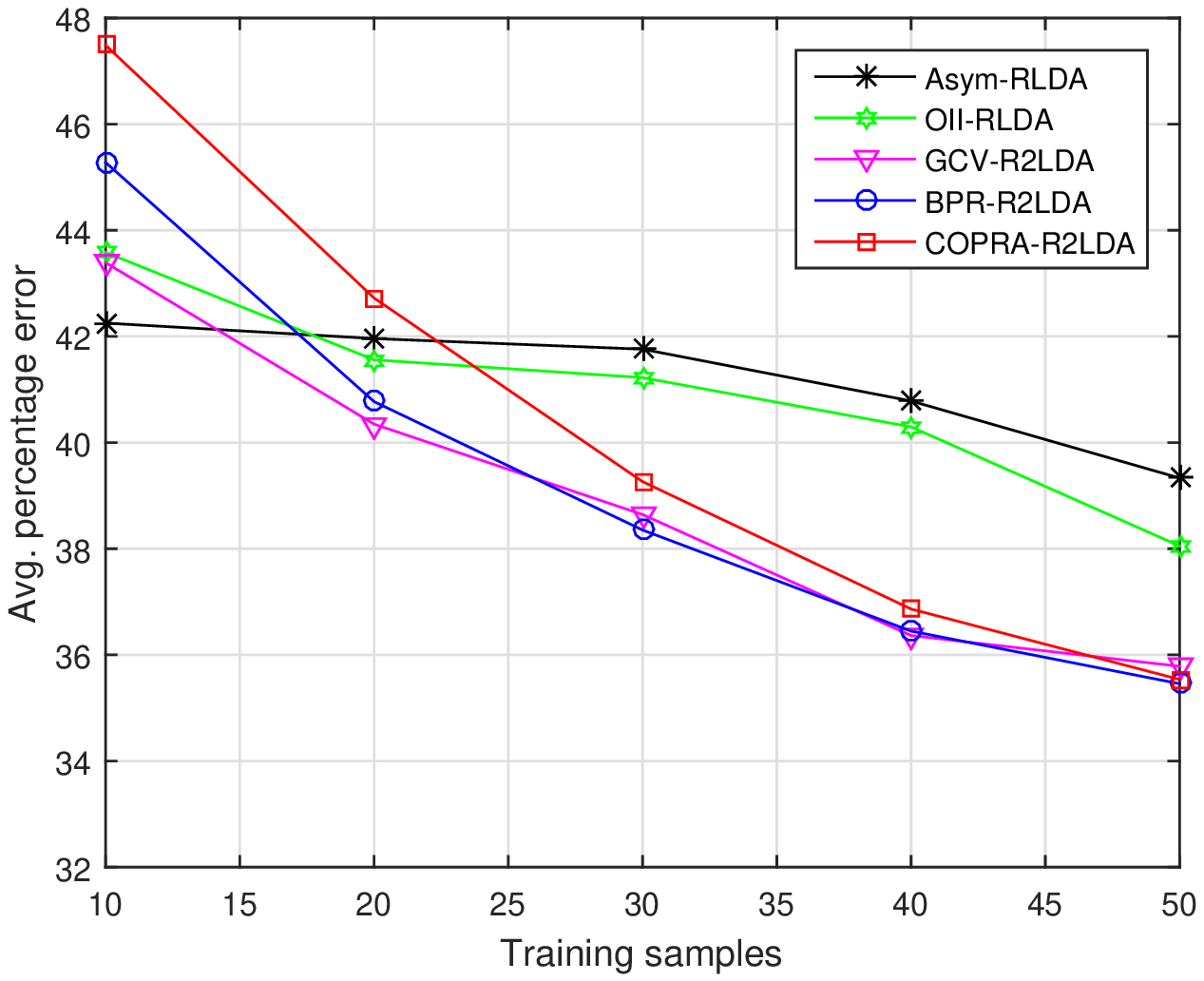}}
\caption{Sonar data misclassification rates versus training data size for different test data noise levels.}
\label{fig:sonar}
\end{center}
\end{figure*}

%\begin{figure*}[!t]
%\begin{center}
%\subfloat[Gaussian, $\sigma=0$]{\includegraphics[width=0.3\textwidth]{GaussRD0.eps}}
%\subfloat[Gaussian, $\sigma=1$]{\includegraphics[width=0.3\textwidth]{GaussRD1.eps%}}
%\subfloat[Gaussian, $\sigma=2$]{\includegraphics[width=0.3\textwidth]{GaussRD2.eps}}
%\caption{Reduced-dimension Gaussian data misclassification rate %versus training data size for different test data noise %levels.}
%\label{fig:GaussDR}
%\end{center}
%\end{figure*}

\begin{figure*}[!t]
\begin{center}
\subfloat[MNIST (1,7), $\sigma = 0$]{\includegraphics[width=0.3\textwidth]{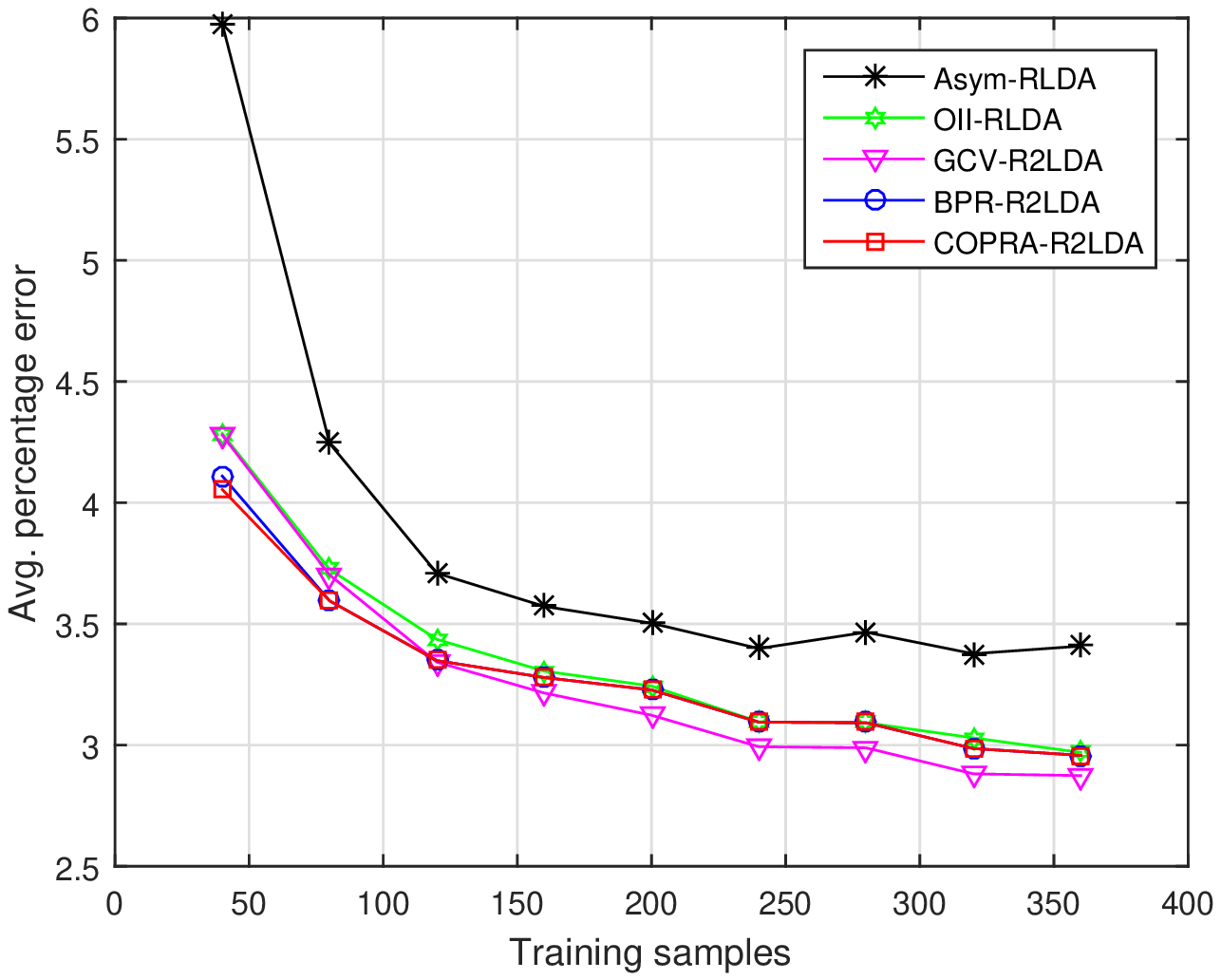}}
\subfloat[MNIST (1,7), $\sigma=1$]{\includegraphics[width=0.3\textwidth]{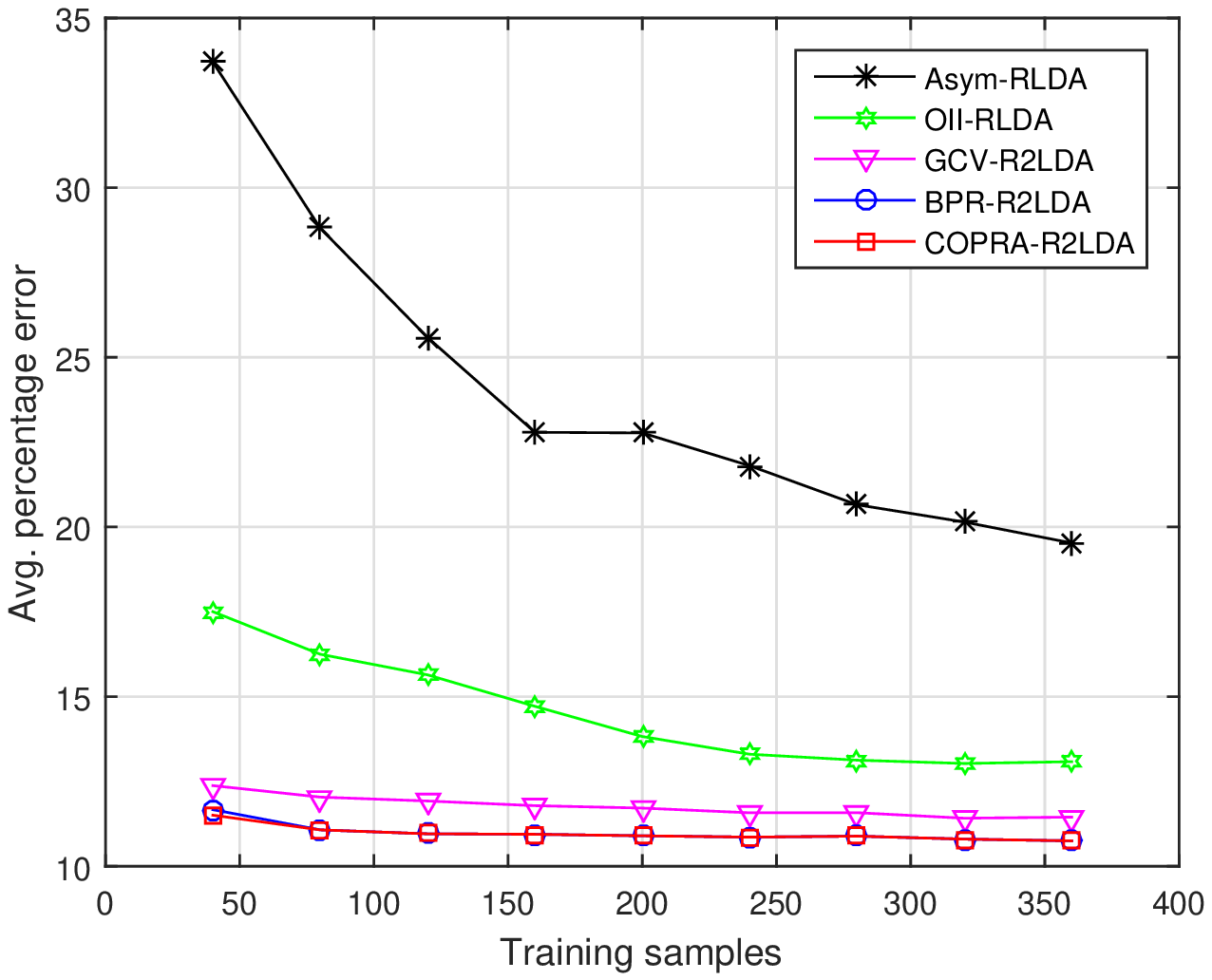}}
\subfloat[MNIST (1,7), $\sigma =2$]{\includegraphics[width=0.3\textwidth]{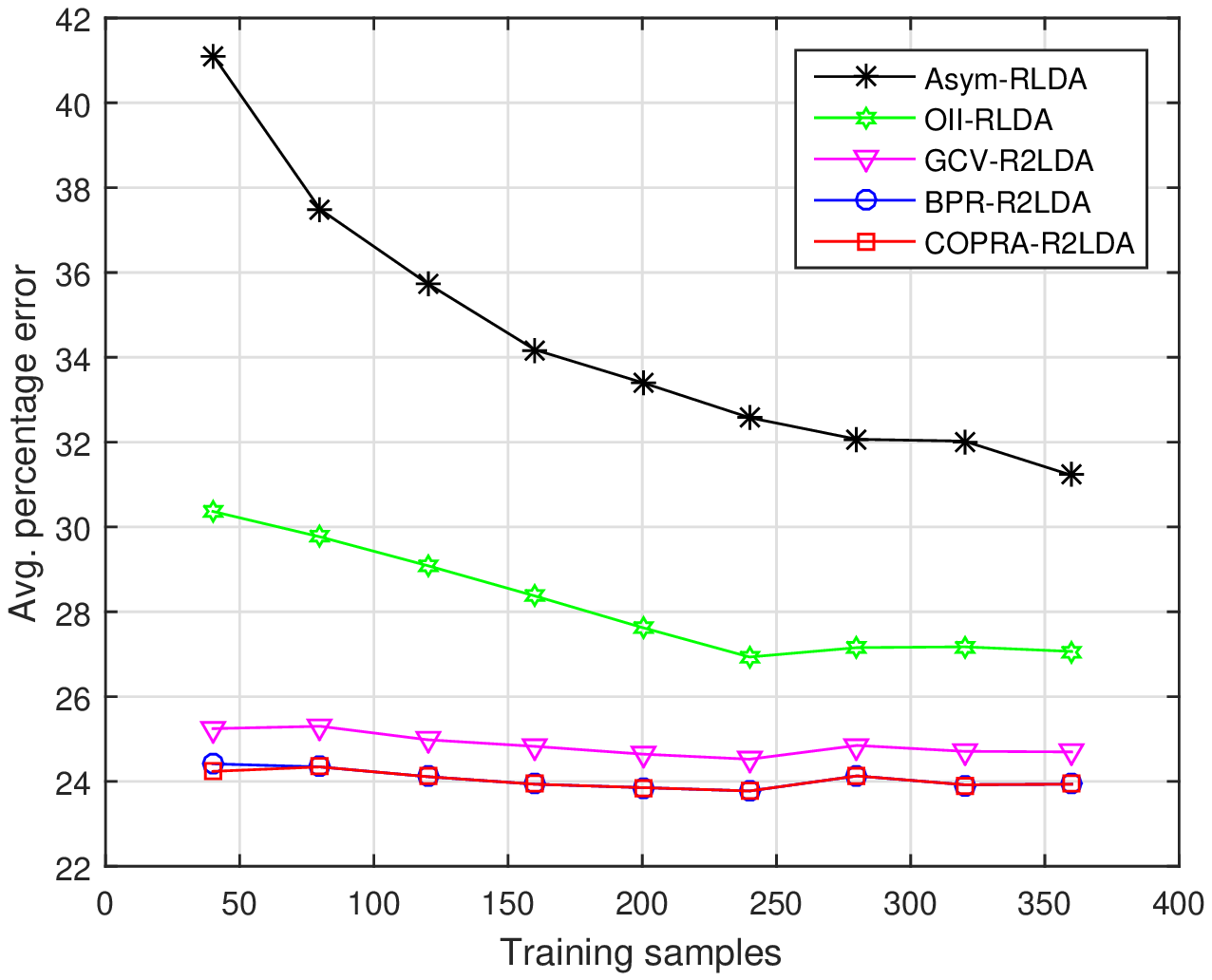}}\\
\subfloat[MNIST (5,8), $\sigma = 0$]{\includegraphics[width=0.3\textwidth]{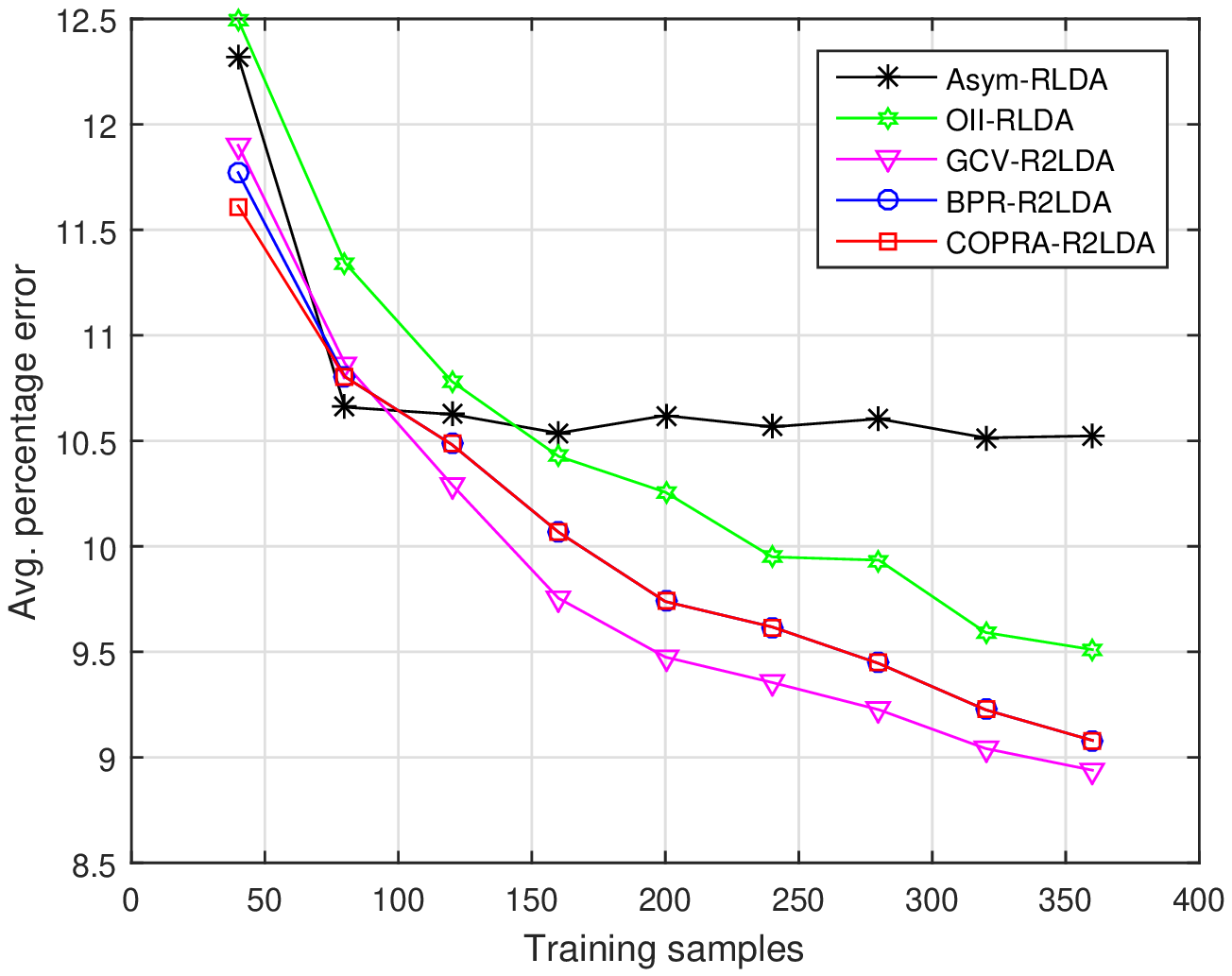}}
\subfloat[MNIST (5,8), $\sigma=1$]{\includegraphics[width=0.3\textwidth]{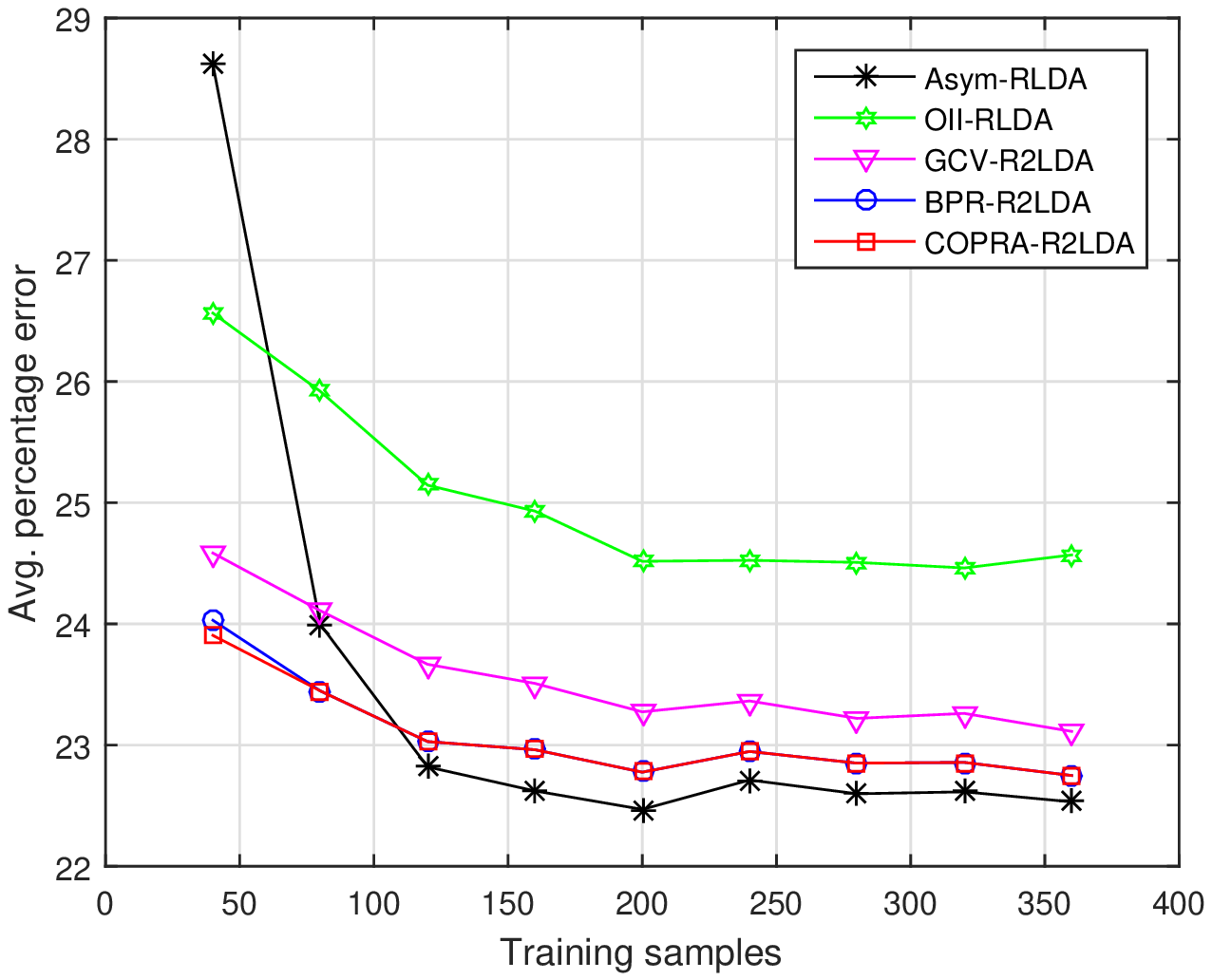}}
\subfloat[MNIST  (5,8), $\sigma =2$]{\includegraphics[width=0.3\textwidth]{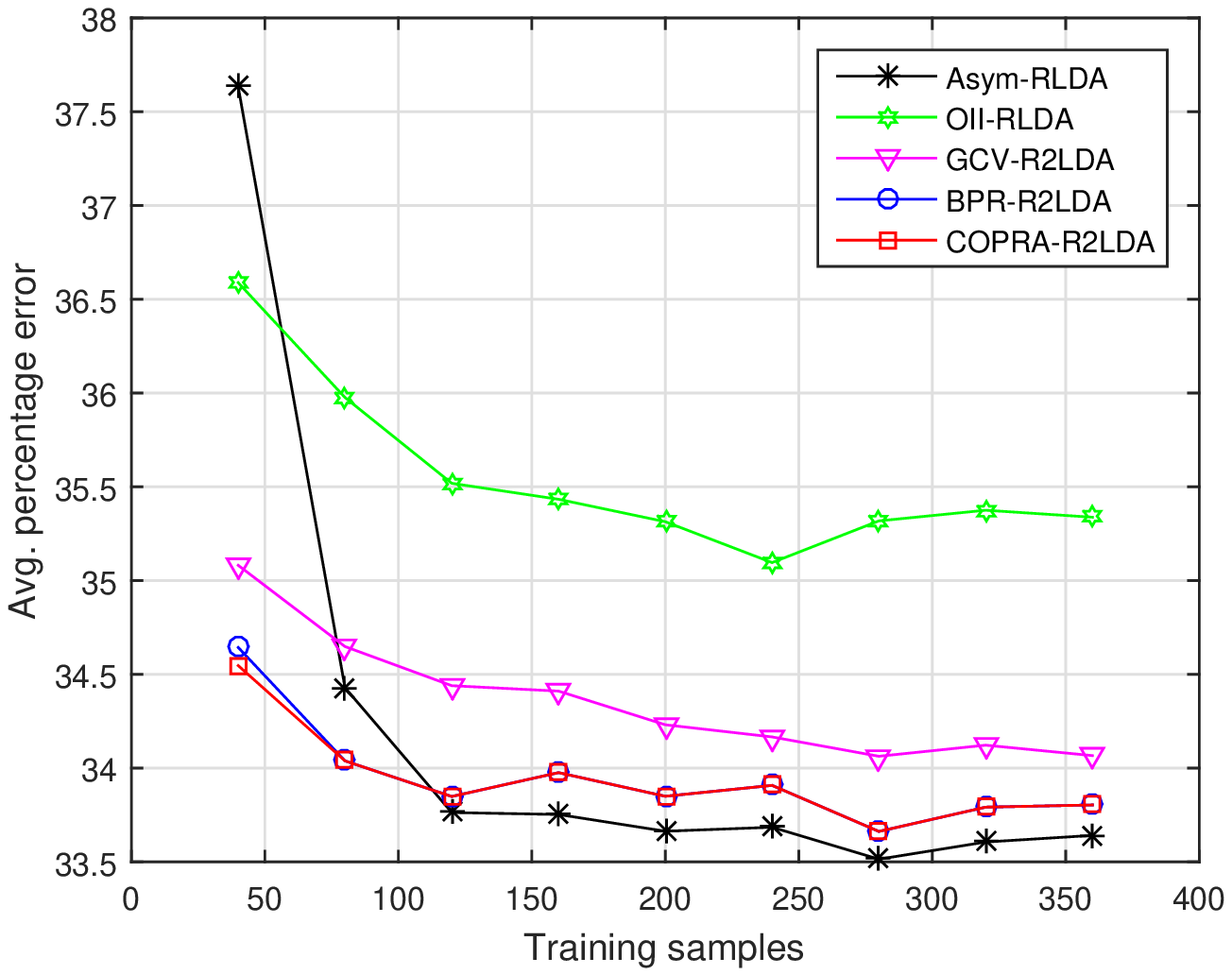}}\\
\subfloat[MNIST (7,9), $\sigma = 0$]{\includegraphics[width=0.3\textwidth]{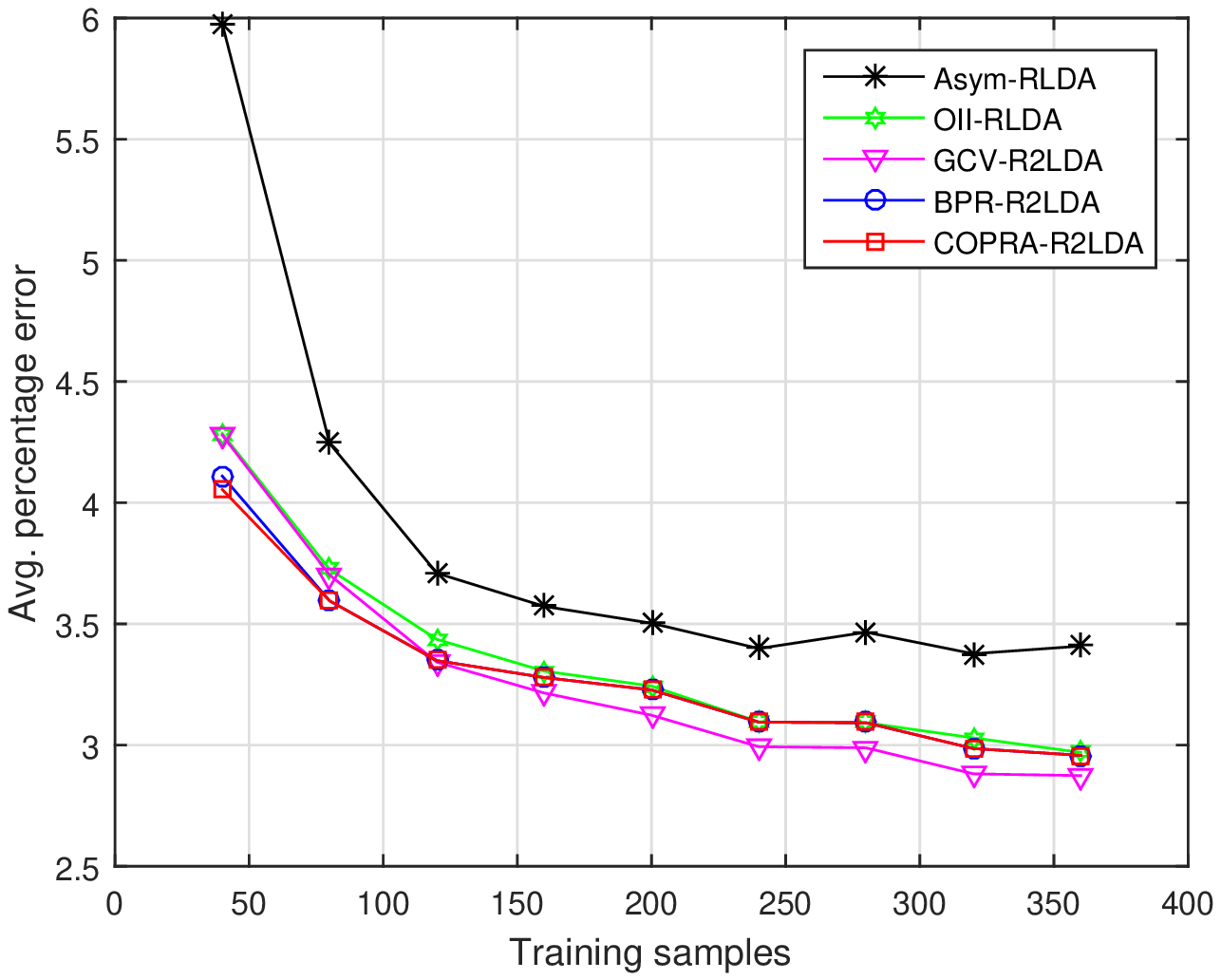}}
\subfloat[MNIST (7,9), $\sigma=1$]{\includegraphics[width=0.3\textwidth]{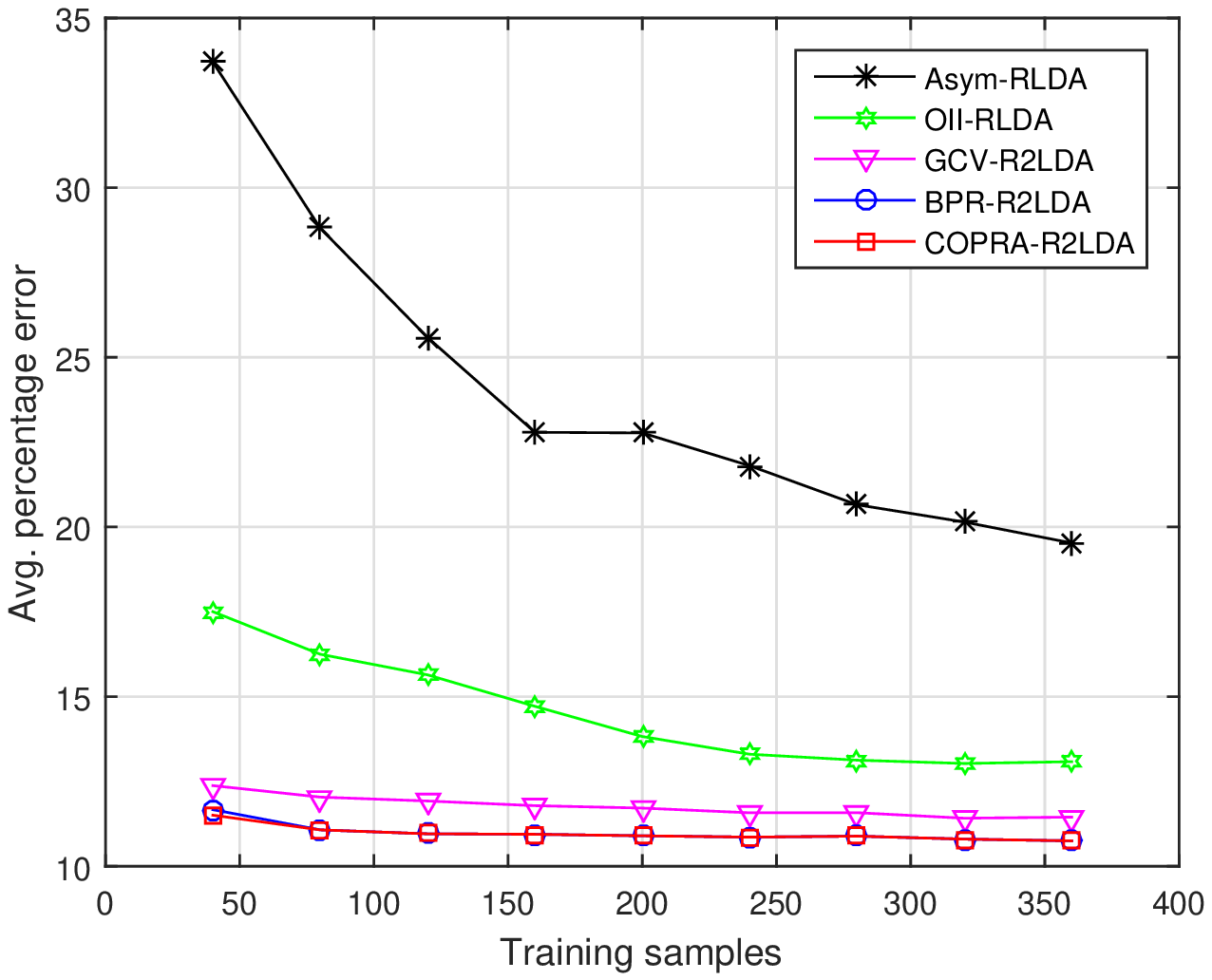}}
\subfloat[MNIST(7,9), $\sigma =2$]{\includegraphics[width=0.3\textwidth]{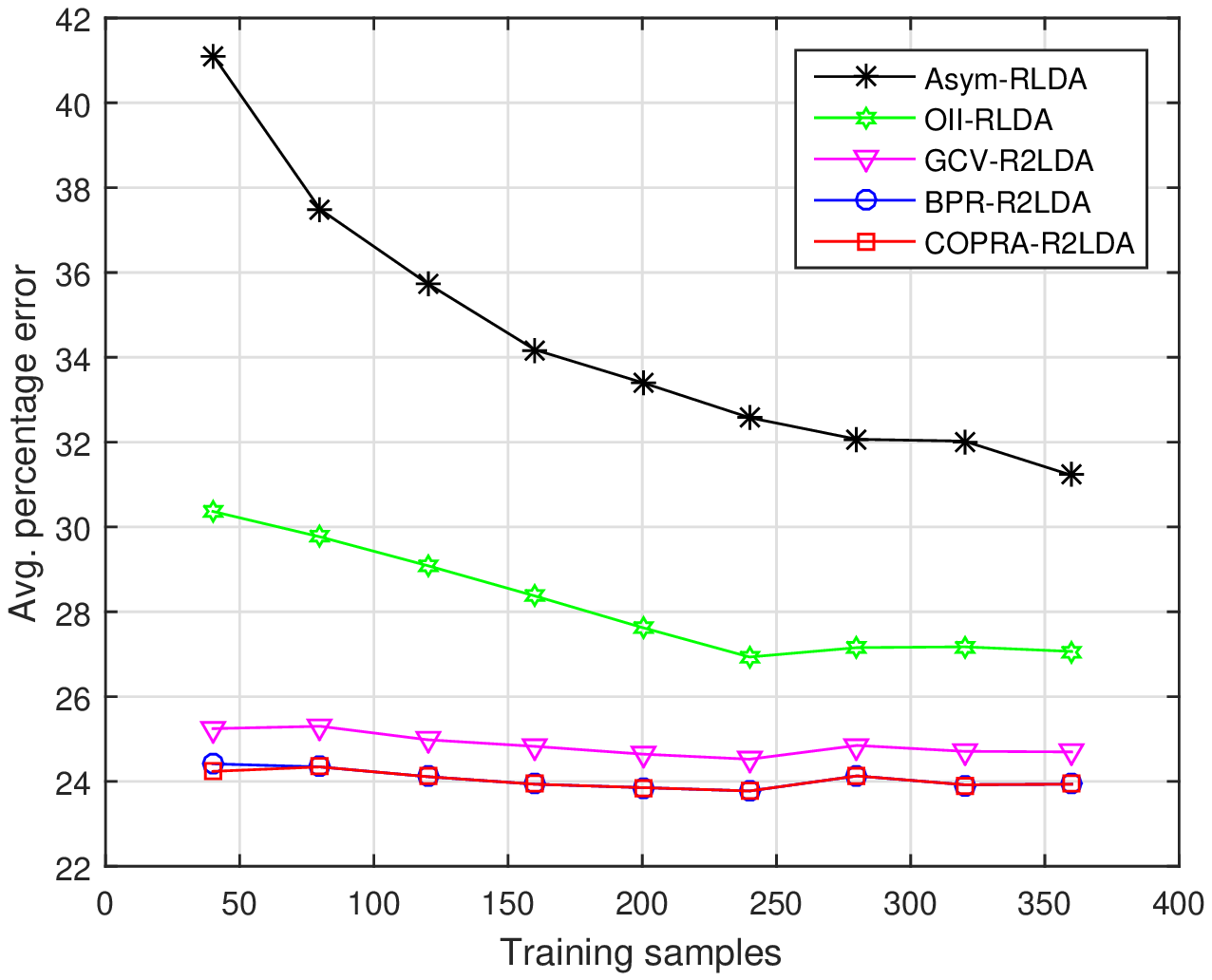}}
\caption{Reduced-dimension MNIST data misclassification rates versus training data size for different test data noise levels.}
\label{fig:MNIST DR}
\end{center}
\end{figure*}

\begin{figure*}[!t]
\begin{center}
\subfloat[One test]{\includegraphics[width=0.45\textwidth]{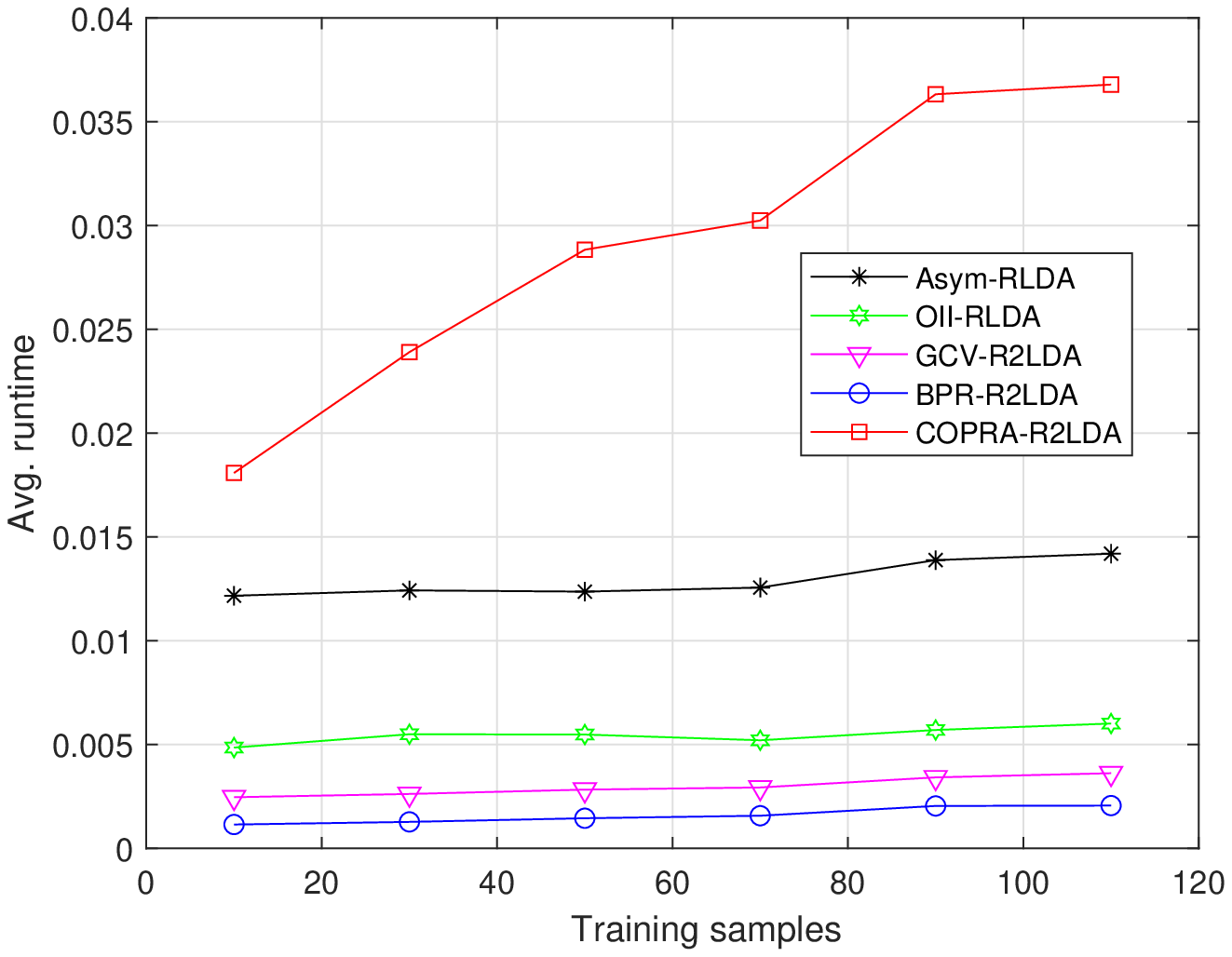}}
\subfloat[500 tests]{\includegraphics[width=0.45\textwidth]{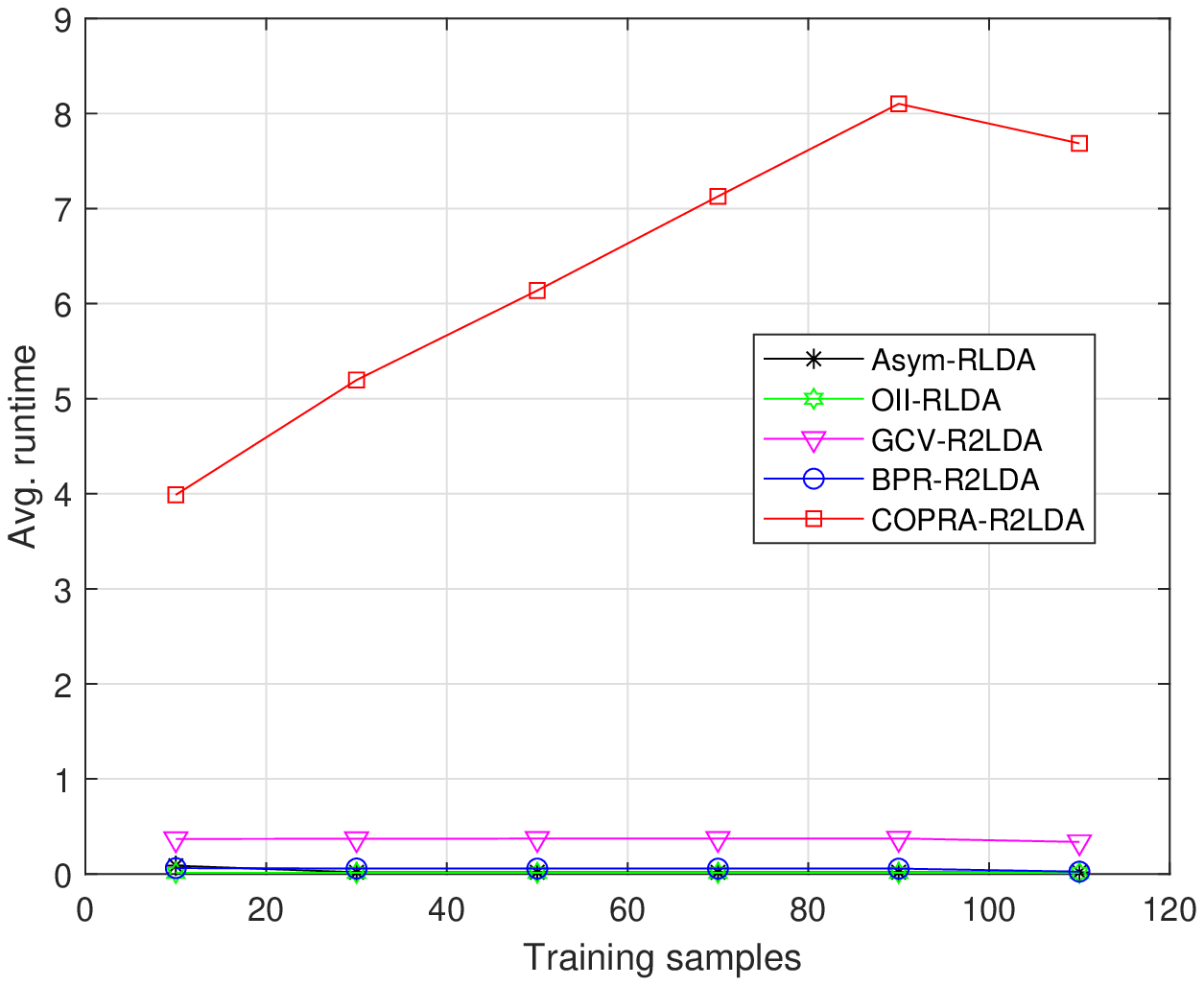}}
\caption{Average runtime (in seconds) versus training data size for Gaussian data: (a) A single test sample, (b) 500 test samples.}
\label{fig:GaussComplexity}
\end{center}
\end{figure*}

\begin{figure*}[!t]
\begin{center}
\subfloat[One test]{\includegraphics[width=0.45\textwidth]{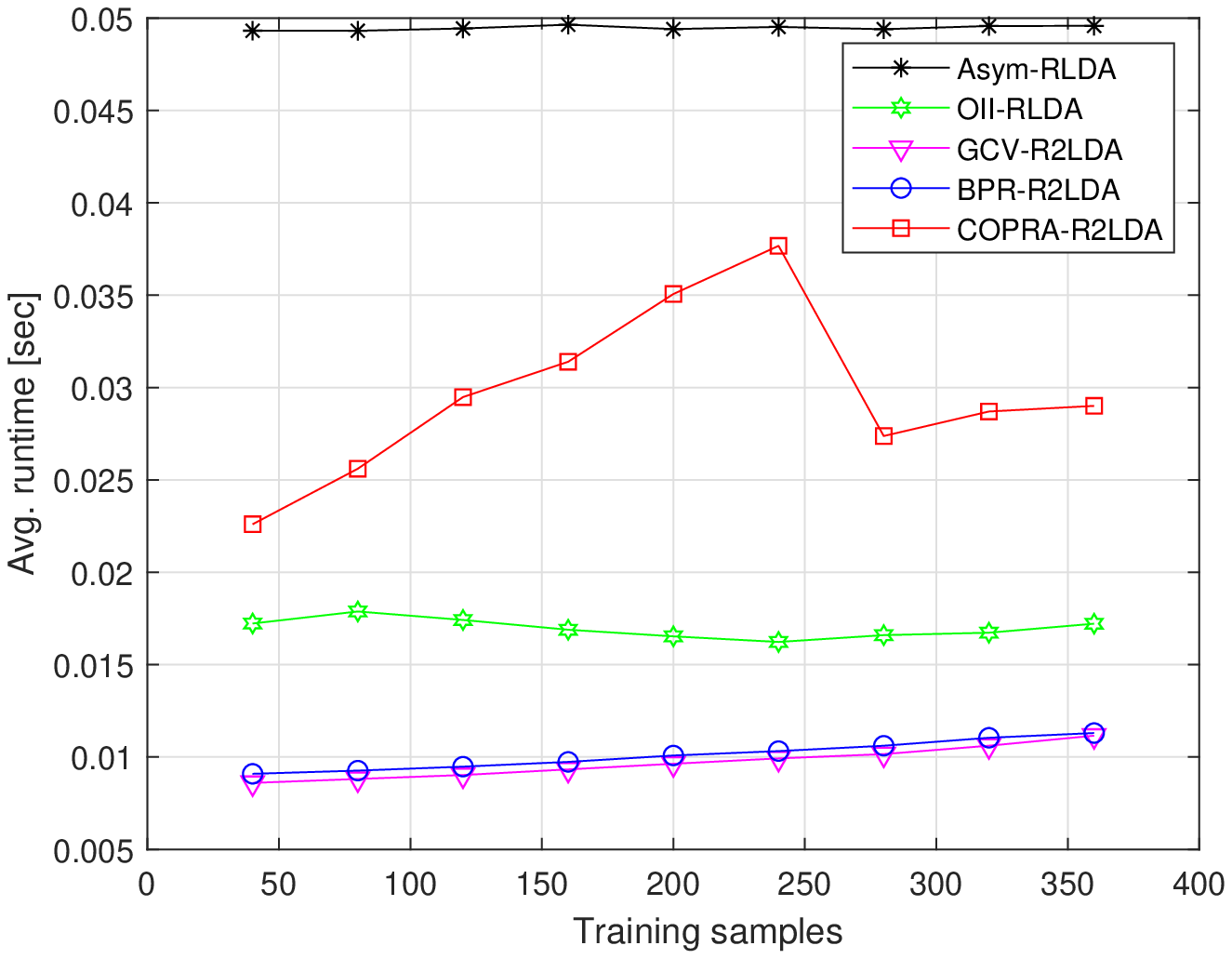}}
\subfloat[500 tests]{\includegraphics[width=0.45\textwidth]{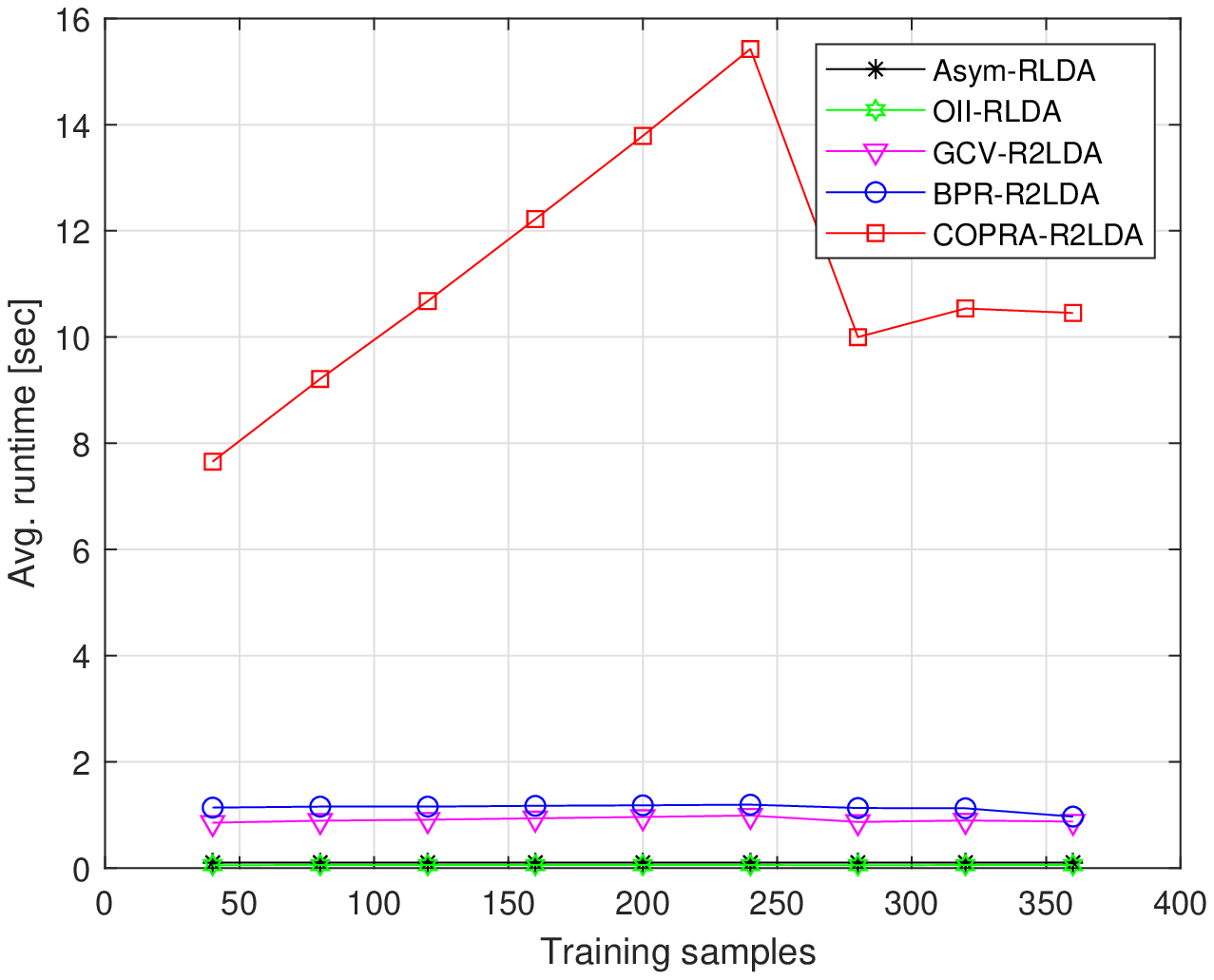}}
\caption{\textcolor{black}{Average runtime (in seconds) versus training data size for the MNIST image pair (7, 9): (a) A single test sample, (b) 500 test samples.}}
\label{fig:MNISTComplexity}
\end{center}
\end{figure*}

%\begin{figure*}[!t]
%\begin{center}
%\subfloat[Gaussian, $\sigma=0$]{\includegraphics[width=0.3\textwidth]{GaussComplexity0_500.eps}}
%\subfloat[Gaussian, $\sigma=1$]{\includegraphics[width=0.3\textwidth]{GaussComplexity1_500.eps}}
%\subfloat[Gaussian, $\sigma=2$]{\includegraphics[width=0.3\textwidth]{GaussComplexity2_500.eps}}
%\caption{500-test average runtime versus training data size for different test data noise levels. }
%\label{fig:GaussComplexity500}
%\end{center}
%\end{figure*}

\subsection{Datasets Description}
\textbf{Synthetic Data:} The synthetic data is generated based on a Gaussian data model with dimension $p = 100$. The class covariance matrix $\Sig_0$ is generated with diagonal elements equal to $1$ and off-diagonal elements equal to $0.1$, while the other class covariance matrix is generated as $\Sig_1\!=\!\Sig_0\!+\!\I$. As for the model mean vectors, we set $\Mu_1 \!=\! -\Mu_0$, where $\Mu_0 \!=\! [a,a,...,a]^{\rm T}$. The parameter $a$ is chosen according to the between-class Mahalanobis distance, $\delta$, defined according to $\delta^2 \!=\! (\Mu_0\!-\!\Mu_1)^{\rm T}\Sig^{-1}(\Mu_0\!-\!\Mu_1)$ \cite{zollanvari2}. We use $\delta^2\!=\!9$. A training set $\mathcal{S}_i$ of size $n_i$ is generated independently in each training trial, where $n_0=n_1$. For the test data, we generate an independent set of samples for each class.

\textbf{Real Data:} We use (i) the MNIST dataset that consists of $20\times20$ gray-scale images of handwritten digits \cite{mnist-dataset}, (ii) the phonemes dataset considered in \cite{phonem2}, and (iii) the sonar classification dataset \cite{sonar}. These datasets are available for download from the UCI Machine Learning Repository \footnote{\url{https://archive.ics.uci.edu/ml/datasets}}.

The MNIST images are vectorized to result in data of dimensionality $p=400$. For binary classification, selected pairs of images are used.

The phonemes dataset is based on log-periodogram (of length $p=256$) of digitized speech frames extracted from the TIMIT database (TIMIT Acoustic-Phonetic Continuous Speech Corpus, NTIS, U.S. Department of Commerce) \cite{phonem2}, which is widely used in speech recognition. The phonemes are transcribed as: (1) \lq\lq sh\rq\rq as in \lq\lq she\rq\rq,  (2) \lq\lq dcl\rq\rq as in \lq\lq dark\rq\rq, (3) \lq\lq iy\rq\rq as the vowel in \lq\lq she\rq\rq, (4) \lq\lq aa\rq\rq as the vowel in \lq\lq dark\rq\rq, and (5) \lq\lq ao\rq\rq as the first vowel in \lq\lq water\rq\rq. For binary classification, selected pairs of phonemes are formed from the above five phonemes.

The sonar dataset consists of 208 examples, each with 60 attributes representing sonar returns from a metal cylinder (class~0) or a rough cylindrical rock (class~1).

\subsection{Experiments Description}
For both the synthetic and real datasets, 500 training trials were carried out, each followed by a number between 50 and 500 test trials, depending on the size of the available of data from the dataset. Each training or test trial is based on a randomly generated/selected data. As a pre-processing step, all datasets are translated to the interval $[-1,1]$ to facilitate comparison of results across different datasets.

For all datasets, we test the case where zero-mean Gaussian noise with standard deviation $\sigma$ is added \emph{only} to the test data. For each dataset, we test $\sigma$ values that allow us to observe reasonable performance variability (some datasets are more resilient to noise than others). The statistical properties of this noise are not known to the proposed R2LDA classifier, nor are they known to any of the benchmark methods.

\subsection{Dimensionality Reduction}
In scenarios involving high-dimensional data and a limited number of observations, one can reduce the dimensionality of the data by extracting a small set of the most significant features present in the data. While there are myriad of feature reduction/selection methods available \cite{fselect}, we apply the simple $t$-test and use the $p$-values of each feature as a criterion for feature selection. In our experiments, we apply dimensionality reduction to the MNIST dataset by selecting the top 12.5\% features based on the $p$-values.
This exercise aims to investigate the behavior of the proposed classifiers in setups with reduced dimensionality.

\subsection{Results Discussion}
Figs.~\ref{fig:Gauss}--\ref{fig:MNIST DR} plot the percentage classification errors versus the training data size ($n$) for different datasets under different test data noise levels. Fig.\ref{fig:Gauss} presents the results for the (synthetic) Gaussian data, while Fig.\ref{fig:MNIST}, Fig.\ref{fig:Phoneme} and Fig.\ref{fig:sonar} show the results for the MNIST, phonemes and sonar datasets, respectively. On the other hand, Fig.~\ref{fig:MNIST DR} depicts results for an example from the MNIST dataset with reduced dimensionality. The MNIST results are based on the image/digit pairs (1,7), (5,8), and (7,9), while the phonemes dataset results use the phoneme combinations (1,2), (1,3), (1,5), and (4,5). From the results in Figs.\ref{fig:Gauss}--\ref{fig:MNIST DR}, we observe the following:
\begin{itemize}
  \item On average, the R2LDA methods outperform the RLDA methods.
  \item The R2LDA methods remain more consistent and stable than the RLDA methods as the noise level in the test data increases. This is more visible in real datasets that deviate from Gaussianity.
  \item Amongst the R2LDA classifiers, COPRA-R2LDA and BPR-R2LDA appear to be slightly more consistent than GCV-R2LDA. GCV-R2LDA seems to occasionally falter, as in Fig.\ref{fig:MNIST}(a), Fig.~\ref{fig:MNIST}(d) and Fig.\ref{fig:MNIST}(g).
  \item For the MNIST dataset with reduced dimensionality, the R2LDA methods preserve their superiority over the RLDA counterparts, especially in noisy conditions. This is evident from Fig.~\ref{fig:MNIST DR}, where the top 50 features are selected out of 400 features present in the MNIST data.
\end{itemize}

\subsection{Computational Complexity}
\textcolor{black}{We consider the computational complexity of the proposed algorithms when classifying a test dataset of size $k$. Let $l_\text{COPRA}$ and $l_\text{BPR}$ be the maximum number of iterations required for the COPRA and BPR algorithms to converge. Also, let $g_\text{GCV}$ and $g_\text{Asym}$ be the number of grid points used in the search processes of the GCV and Asym methods, respectively. The worst-case time complexities of the proposed algorithms (including all the steps listed in Subsection~\ref{subsec:rrlda_algo}) and the benchmark methods are given in Table~\ref{tab:comp} using the big-O notation.}
\begin{table}[!t]
    \centering
    \renewcommand{\arraystretch}{1.5}
    \caption{Time complexity summary}
    \label{tab:comp}
    \begin{tabular}{|c|l|l|}
    \hline
    \textbf{No.} &
    \textbf{Algorithm} &  \textbf{Complexity} \\ \hline
    1 & Asym-RLDA & $\mathcal{O} (n p^2 + p^3 + k p^2 + g_\text{Asym} p^2)$ \\ \hline
    2 & OII-RLDA & $\mathcal{O} (n p^2 + p^3 + k p^2)$ \\ \hline
    3 & COPRA-R2LDA & $\mathcal{O} (n p^2 + p^3 + l_\text{COPRA} k p^2)$ \\ \hline
    4 & BPR-R2LDA & $\mathcal{O} (n p^2 + p^3 + l_\text{BPR} k p^2)$ \\ \hline
     5 & GCV-R2LDA & $\mathcal{O} (n p^2 + p^3 + g_\text{GCV} k p^2)$ \\ \hline
    \end{tabular}
\end{table}
%\begin{itemize}
%  \item COPRA-R2LDA is $\mathcal{O} (n p^2 + p^3 + l_\text{COPRA} k p^2)$,
%  \item BPR-R2LDA is $\mathcal{O} (n p^2 + p^3 + l_\text{BPR} k p^2)$, and
%  \item GCV-R2LDA is $\mathcal{O} (n p^2 + p^3 + g_\text{GCV} k p^2)$.
%\end{itemize}
%Likewise, for the benchmark methods, we can establish that
%\begin{itemize}
%  \item Asym-R2LDA is $\mathcal{O} (n p^2 + p^3 + k p^2 + g_\text{Asym} p^2)$, and
%  \item OII-R2LDA is $\mathcal{O} (n p^2 + p^3 + k p^2)$.
%\end{itemize}

\textcolor{black}{Note that all the five complexity expressions listed in Table~\ref{tab:comp} feature the terms $n p^2$ and $p^3$. These two terms are, approximately, of similar order for scenarios with $n\approx p$. Each complexity expression includes a term of the form $\alpha k p^2$, with different $\alpha$ values for different methods. For a large $\alpha$ and/or a large number of test samples $k\gg p$, this term will dominate the complexity.
For the RLDA methods, we have $\alpha=1$. On the other hand, for the R2LDA methods, $\alpha$ takes the values $l_\text{COPRA}$ and $l_\text{BPR}$ and $g_\text{GCV}$, for the three methods respectively. These parameters are due to the computations involved in finding the regularization parameter $\gamma_z$ each time a test data sample is classified. As an example, for $n\approx k\approx p$, an R2LDA algorithm with $\alpha \approx p$ would have a complexity $\mathcal{O} (p^4)$. Under the same conditions, an RLDA algorithm's complexity is $\mathcal{O} (p^3)$.
}

\textcolor{black}{In addition to the time complexity, we also consider the runtimes of various algorithms observed during our experiments. We illustrate this using two examples.}
Fig.~\ref{fig:GaussComplexity} compares the runtimes (in seconds) of various algorithms against the number of training samples for the Gaussian data used in Fig.~\ref{fig:Gauss}. Fig.~\ref{fig:GaussComplexity}(a) and Fig.~\ref{fig:GaussComplexity}(b) plot the average runtime for a single test sample and 500 test samples, respectively. We observe that the COPRA-R2LDA is considerably slower than the other algorithms for both numbers of test data samples. Despite computing a new regularization parameter for each test data sample, BPR-R2LDA and GCV-R2LDA offer comparable runtimes to those of the benchmark RLDA methods.

\textcolor{black}{In Fig.~\ref{fig:MNISTComplexity}, we show another example similar to Fig.~\ref{fig:GaussComplexity} using the MNIST dataset. In this example, COPRA-R2LDA is faster than Asym-RLDA in the single-test case. Whereas, with 500 tests, COPRA-R2LDA becomes substantially slower than the rest of the algorithms. On the other hand, the runtimes of BPR-R2LDA and GCV-R2LDA stay relatively close to those of the RLDA methods when applied to 500 test samples, while offering the fastest runtimes in the single-test case. The slowness of the COPRA-R2LDA algorithm is attributed mainly to its large convergence time.}

\textcolor{black}{Based on the above discussions, we can conclude that, among the tested algorithms, BPR-R2LDA is the most attractive classifier since it is much faster than COPRA-R2LDA and offers a more consistent classification performance than GCV-R2LDA.}

\section{Conclusions}\label{sec:concl}
We have presented novel regularized LDA classifiers based on a dual regularization scheme. The proposed R2LDA approach allows us to tune two regularization parameters independently. The first regularization parameter is computed offline from the training data. In contrast, the second regularization parameter is dynamically tuned to each test data sample. Based on synthetic and real datasets, results confirm our approach's effectiveness. The results also demonstrate the robustness of the proposed approach when noise is present in the test data. Although the proposed method is developed for binary classification, it can be easily extended to the multi-class case.
% -------------------------------------------------------------------------
\newpage
\bibliographystyle{IEEEtran}
\bibliography{refs}

\end{document}